\documentclass{article}

\usepackage[preprint]{neurips_2024}

\usepackage[utf8]{inputenc} 
\usepackage[T1]{fontenc}    
\usepackage{hyperref}       
\usepackage{url}            
\usepackage{booktabs}       
\usepackage{amsfonts}       
\usepackage{nicefrac}       
\usepackage{microtype}      
\usepackage[dvipsnames]{xcolor}         
\usepackage{wrapfig}
\usepackage{lipsum}
\usepackage[most]{tcolorbox}

\usepackage{multicol}
\usepackage{multirow}

\usepackage{listings} 
\usepackage{color} 

\definecolor{codegray}{rgb}{0.5,0.5,0.5}
\definecolor{codegreen}{rgb}{0,0.6,0}
\definecolor{codepurple}{rgb}{0.58,0,0.82}
\definecolor{backcolour}{rgb}{0.95,0.95,0.92}

\lstdefinestyle{mystyle}{
    backgroundcolor=\color{backcolour},   
    commentstyle=\color{codegreen},
    keywordstyle=\color{magenta},
    numberstyle=\tiny\color{codegray},
    stringstyle=\color{codepurple},
    basicstyle=\ttfamily\footnotesize,
    breakatwhitespace=false,         
    breaklines=true,                 
    captionpos=b,                    
    keepspaces=true,                 
    numbers=left,                    
    numbersep=5pt,                  
    showspaces=false,                
    showstringspaces=false,
    showtabs=false,                  
    tabsize=2
}
\usepackage{graphicx}
\usepackage{amsmath}

\title{Can LLMs Serve As Time Series Anomaly Detectors?}

\author{%
  Manqing Dong\textsuperscript{\dag}, Hao Huang\textsuperscript{\ddag}, and Longbing Cao\textsuperscript{\dag}
    \\
  \textsuperscript{\dag}School of Computing, Macquarie University\\
  \textsuperscript{\ddag}School of Computer Science, University of Technology Sydney\\
}

\begin{document}

\maketitle

\begin{abstract}
An emerging topic in large language models (LLMs) is their application to time series forecasting, characterizing mainstream and patternable characteristics of time series. A relevant but rarely explored and more challenging question is whether LLMs can detect and explain time series anomalies, a critical task across various real-world applications. In this paper, we investigate the capabilities of LLMs, specifically GPT-4 and LLaMA3, in detecting and explaining anomalies in time series. Our studies reveal that: 1) LLMs cannot be directly used for time series anomaly detection. 2) By designing prompt strategies such as in-context learning and chain-of-thought prompting, GPT-4 can detect time series anomalies with results competitive to baseline methods. 3) We propose a synthesized dataset to automatically generate time series anomalies with corresponding explanations. By applying instruction fine-tuning on this dataset, LLaMA3 demonstrates improved performance in time series anomaly detection tasks. In summary, our exploration shows the promising potential of LLMs as time series anomaly detectors. 
\end{abstract}

\section{Introduction}

With the  capabilities of Large Language Models (LLMs) demonstrated in handling various tasks, particularly for natural language processing (NLP)~\citep{achiam2023gpt} and computer vision (CV)~\citep{liu2024visual}, LLMs-based time series analysis emerges as a promising topic~\citep{zhang2024large}. Their primary focus is on time series forecasting, which is an increasingly concerned topic for its broad and lasting roles in wide applications. Their studies can be broadly classified into two groups: (1) prompt engineering approaches, where time series are treated as a series of tokens, either directly fed into the LLMs~\citep{gruver2023large} or combined with instruction prompts~\citep{xue2023promptcast}, to conduct time series forecasting in a sentence-to-sentence fashion; and (2) aligning approaches, which use LLMs as backbones to train encoders transforming time series into embeddings and decoders  translating the LLM outputs into the required output, or even utilizing the middle layers of the LLMs via strategies like pretraining~\citep{ansari2024chronos} or parameter-efficient fine-tuning (PEFT)~\citep{he2022towards,zhou2023one,jin2023time}. In contrast, time series anomaly detection, while increasingly studied in deep anomaly detection \citep{Pang-odr21}, has been rarely explored in the realm of LLMs. 

LLMs-based time series anomaly detection exhibits significant challenges differing from LLMs-based time series forecasting. The latter captures mainstream and patternable characteristics in time series, while the former needs to handle anomaly complexities including point and contextual exceptions. The limited work available on LLMs for time series anomaly detection \citep{zhou2023one,zhang2023large,liu2024large} does not explicitly verify or address these issues. They also overlook the textual reasoning ability of LLMs, treating both the input and output of LLMs as time series, without the explanation of how LLMs make their decisions. This motivates us to investigate an important capability area of LLMs in this paper: can LLMs serve as explainable time series anomaly detectors?

First, inspired by the work on treating LLMs as zero-shot learners for time series forecasting through prompt engineering~\citep{gruver2023large,liu2024lstprompt}, we investigate whether LLMs can understand general anomaly-sensitive patterns in time series and explain their decisions, which are essential for time series anomaly detection tasks. Second, time series anomalies can present in different forms, such as point anomalies and contextual anomalies. Therefore, we explore not only whether LLMs can detect time series anomalies but also whether they can identify specific types of anomalies. This approach goes beyond treating time series anomaly detection as a binary classification task~\citep{zhou2023one,zhang2023large}.

Specifically, we investigate the ability of two representative LLMs, GPT-4~\citep{achiam2023gpt} and LLaMA3\footnote{\url{https://llama.meta.com/llama3/}}, for time series anomaly detection and their explainability by addressing three questions: (1) \textit{Can LLMs be directly applied for explainable time series anomaly detection?} Unfortunately, the answer is no, leading us to the next question: (2) \textit{How can LLMs detect and explain time series anomalies via designing appropriate prompt strategies?} Through various tests, we find that GPT-4 often excels as an explainable time series anomaly detector with minimal prompt instructions. However, our study also reveals gaps in the performance of smaller LLMs, i.e., LLaMA3, in prompt-based time series anomaly detection. This brings us to our final question: (3) \textit{Can we improve LLMs' detection performance by designing proper instruction fine-tuning?} Since there is no available time series data with both anomalies and explanations for instruction fine-tuning, we propose a Time Series and Text Explanation Generator (TTGenerator) to automatically generate time series with anomalies and corresponding descriptions for base and anomaly patterns. We demonstrate the augmentation and benchmarking roles of this dataset in enhancing LLM-based time series anomaly detection.

In a nutshell, our contributions include:
\begin{itemize}
\item Comprehensively investigating the zero-shot learning performance of LLMs in time series anomaly detection tasks and their explanatory capabilities.
\item Proposing strategic prompt engineering enabling advanced LLMs to achieve competent performance in anomaly detection, compared to baseline methods.
\item Introducing a synthesized dataset for fine-tuning LLMs, enhancing their performance in time series anomaly detection tasks post fine-tuning.
\end{itemize}
To the best of our knowledge, this work represents a very first to comprehensively investigate and enhance the performance of LLMs in time series anomaly detection, with specific strategies designed to expand LLMs to the broad time series and anomaly detection domains.

\section{Related Work}
Transformers have demonstrated remarkable success in natural language processing (NLP) and, given their proficiency in handling sequential data, a significant number of transformer-based models have been proposed for time series forecasting. Early works focused on modifications to transformer modules, ranging from position embeddings~\citep{nie2022time} to attention mechanisms~\citep{zhou2021informer}, to better fit time series analysis~\citep{zhou2023one}. Most approaches can be regarded as aligning approaches, as they start training from transformer backbones (such as BERT~\citep{kenton2019bert}, GPT-2~\citep{radford2019language}, and T5~\citep{raffel2020exploring}), and train the encoder, decoder, or middle layers of the transformers via fine-tuning~\citep{zhou2023one,chang2024llm4ts,cao2024tempo} or full parameter pretraining~\citep{ansari2024chronos}. However, these models often overlook the rich textual information within the pretrained models, where the fine-tuned model is still primarily used to process only time series data.

Starting with ChatGPT~\citep{ouyang2022training}, we have witnessed the power of large language models (LLMs) such as GPTs~\citep{achiam2023gpt} and LLaMAs~\citep{touvron2023llama,touvron2023llama2}. These models, with larger parameters and trained on more extensive datasets~\citep{hoffmann2022training}, exhibit powerful reasoning capabilities for handling complex tasks. This has triggered initial research in time series analysis, where some studies directly treat the time series as tokens and feed them into LLMs for forecasting~\citep{gruver2023large}, or incorporate time series data with instruction prompts~\citep{xue2023promptcast} or chain-of-thought prompts~\citep{liu2024lstprompt}. Some even further fine-tune the LLMs~\citep{jin2023time}. However, these outputs remain within the time series domain, limiting their applicability to tasks such as generating descriptions for time series data.

Few works have investigated utilizing LLMs for time series anomaly detection, a critical task in various real-world applications. \cite{zhou2023one} is likely the first to fine-tune language models for time series anomaly detection. They fine-tuned a general model for diverse time series tasks, such as classification, anomaly detection, forecasting, and few-shot or zero-shot learning, treating anomaly detection as a binary classification problem and adding layers on top of transformer modules for classification. \cite{liu2024large} used LLMs as the teacher model and trained a student network to mimic the LLM outputs, identifying anomalies as points with distinct values between the teacher and student networks. However, these works did not utilize the reasoning ability of LLMs to provide textual explanations for detections.
\cite{zhang2023large} evaluated GPT-4 and Claude-2~\footnote{\url{https://www.anthropic.com/claude}} with prompt engineering for human mobility trajectory behavior anomaly detection, asking the models to provide explanations for the detections. However, they treated anomaly detection as a binary classification task (i.e., whether a given human mobility trajectory behavior sequence contains anomalies) and did not delve into the explanations provided by the LLMs.
In this work, we focus on detecting specific anomaly points or segments within a time series and investigate whether LLMs can accurately capture the indices of anomalies and explain their detection results. To the best of our knowledge, this is one of the first comprehensive studies on time series anomaly detection using LLMs.

\color{black}
\begin{figure}[t]
    \centering
    \includegraphics[width=0.9\linewidth]{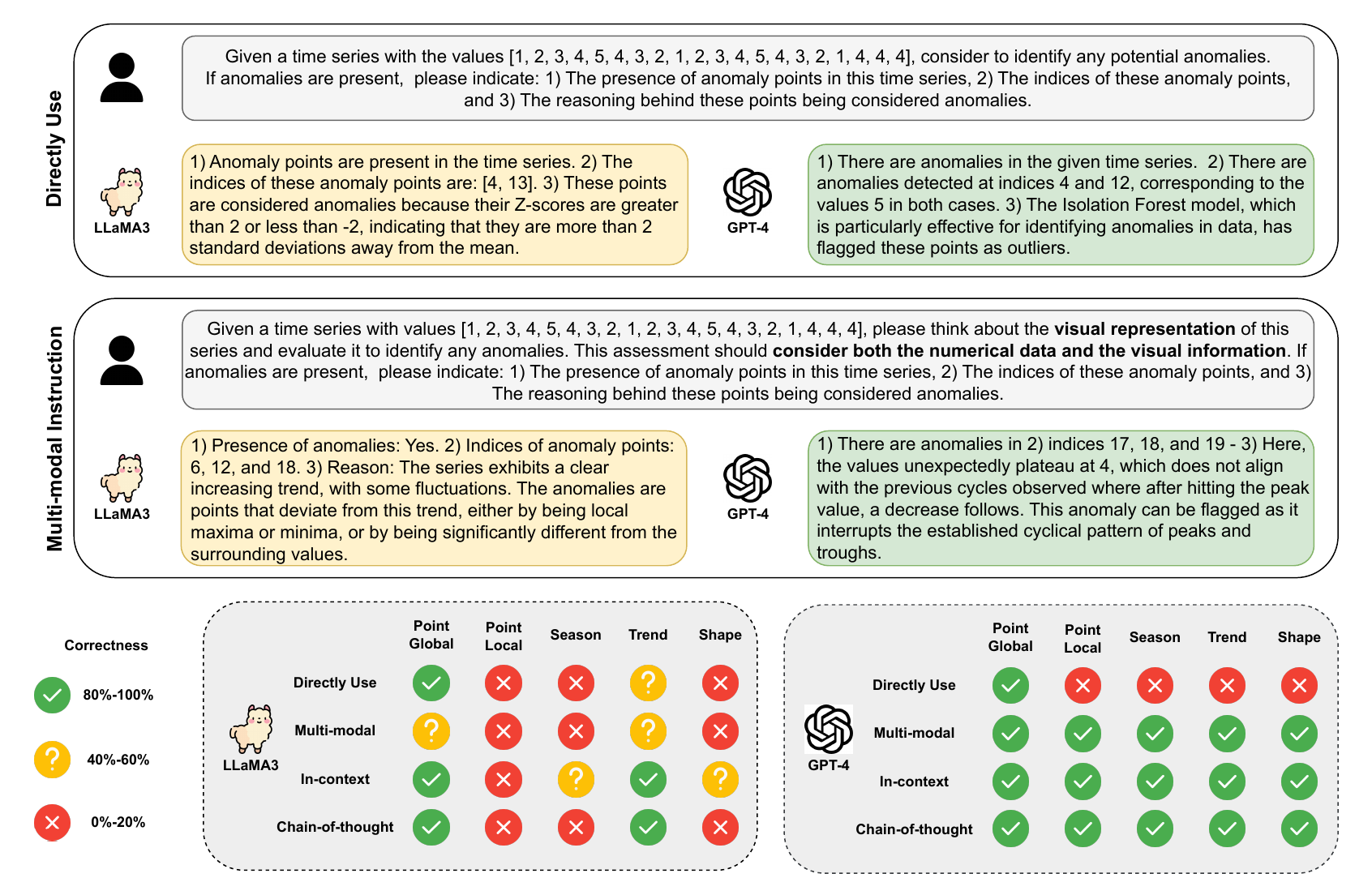}
    \caption{Example of responses from LLaMA-3 and GPT-4 to time series with shape anomalies using direct and multi-modal instructions. The bottom panel shows the overall performance across all trial examples for different anomaly types: global point anomalies, local point anomalies, seasonal anomalies, trend anomalies, and shape anomalies; and prompting strategies: directly use LLMs, multimodal instruction, in-context and chain-of-thought strategies anomaly detection. For each anomaly type and prompting strategy, we conduct five trials and evaluate the correctness of both identified indices and explanations. A correctness rate of 100\% means the model provided correct results in all five trials.}
    \label{fig:example_response}
\end{figure}

\section{Can LLMs Be Directly Applied for Time Series Anomaly Detection? }
\label{sec:empirical_study}
We begin with an empirical study on evaluating two representative large language models (LLMs), GPT-4 and LLaMA-3, in identifying and explaining anomalies in time series data. Our approach involves interpreting time series data as text tokens and tasking the LLMs with: i) determining the presence of anomalies, and ii) if anomalies are identified, providing the indices for the anomalies and explaining the reasons.
We assess the LLMs in terms of their capabilities in detecting five representative types of time series anomalies: global point anomaly, local point anomaly, seasonality anomaly, trend anomaly, and shape anomaly. Detailed descriptions of these anomalies can be found in Section~\ref{sec:ttgenerator}, and examples of each anomaly type are provided in Appendix~\ref{app:trial_examples}.

In Figure~\ref{fig:example_response}, the top part illustrates the performance of these LLMs on a short time series with shape anomalies at indices 17, 18, and 19. Unfortunately, after five trials, neither model achieves accurate results. Similar outcomes are observed for other anomaly types, including local point anomalies, seasonality anomalies, and trend anomalies. However, both LLaMA-3 and GPT-4 perform well in detecting global point anomalies. This indicates that LLMs cannot be directly applied to detect most typical time series anomalies.

Upon examining the intermediate reasoning steps of these LLMs, it appears that they involve simplistic methodologies, such as Isolation Forest~\citep{liu2008isolation} and the z-score technique, as shown in the figure. These approaches make it easier to identify global point anomalies. Unlike GPT-4, which may leverage external tools including Python, LLaMA-3's responses are solely derived from its textual reasoning capabilities. This occasionally results in hallucinated calculations and indices in its responses. Despite this, LLaMA-3 seems to intuitively understand the indices and corresponding values in the time series, particularly in the example shown in the figure, recognizing the significance of indices 4 and 12 for the value 5, even though these are not the actual anomalies.

In summary, the operational logic of these LLMs for time series anomaly detection can be characterized as follows: they first select a suitable anomaly detection strategy, identify the time series sequences within the input, and then construct their responses based on this strategy. However, based on our exploration, we cannot directly apply LLMs for time series anomaly detection, in particular, comprehensive anomaly types in time series.

\section{How to Make LLMs An Explainable Time Series Anomaly Detector via Prompt Engineering?}
\label{sec:prompt_evaluation}

\begin{figure}[t]
    \centering
    \vspace{-10pt}
    \includegraphics[width=\linewidth]{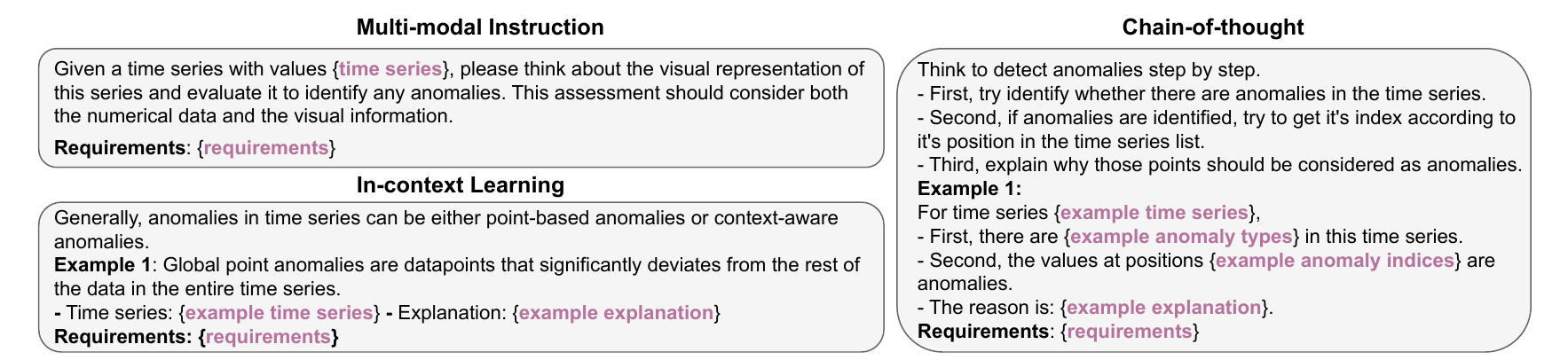}
    \caption{Templates for different prompt strategies, where the `\texttt{requirements}' include the tasks for the LLMs to do, e.g., providing the indices for the anomalies, and explaining the reason if anomalies are detected, with examples in Figure~\ref{fig:example_response}. More details can be found in Appendix~\ref{app:prompt_settings}}
    \label{fig:example_prompt_strategies}
    \vspace{-12pt}
\end{figure}


\subsection{Prompting Strategies}

\textbf{Multi-modal Instruction}
Since LLMs seem to grasp the overall shape of time series, we add prompts to guide the LLMs to also consider the visual representation of the time series for anomaly detection. Refer to the example prompt template in Figure~\ref{fig:example_prompt_strategies}.

\textbf{In-context Learning~\citep{dong2023survey}}
In-context learning is a common prompting approach that includes $n$-shot examples in the prompts to help LLMs with target tasks. For time series anomaly detection, we include examples of five anomaly types: global point anomalies, local point anomalies, seasonality anomalies, trend anomalies, and shape anomalies, respectively. More details about these types of anomalies can be found in Section~\ref{sec:ttgenerator}. Figure~\ref{fig:example_prompt_strategies} shows an example of 1-shot in-context learning with an example of global point anomalies. This includes a brief description of the characteristics of global point anomalies, an example time series containing global point anomalies, and an example of the desired explanation for this time series anomaly detection.

\textbf{Chain-of-thought Prompting~\citep{wei2022chain}}
Chain-of-thought prompting further guides LLMs to decompose complex questions into detailed intermediate reasoning steps. For time series anomaly detection, humans typically first look at the whole time series to detect whether there are anomalies. If anomalies are detected, they then examine each anomaly in detail and explain the reasons for being anomalies. We formalize this process into a prompt, as illustrated in Figure~\ref{fig:example_prompt_strategies}. The simplest approach is to directly request the model to follow these reasoning steps. Alternatively, we can also add $n$-shot examples for the reasoning process.

\subsection{Performance on Trial Examples}
We apply the above prompting strategies to design the trial examples for the five types of anomalies and evaluate the ability of the LLMs in identifying and explaining such anomalies in time series. The bottom part in Figure~\ref{fig:example_response} shows the overall performance of LLaMA-3 and GPT-4 on these trial examples. Additional examples of the responses from LLaMA-3 and GPT-4 can be found in Appendix~\ref{app:trial_examples}. 

Generally, we observe that LLaMA-3 does not show significant improvement with different prompt designs and example cases, while GPT-4 demonstrates impressive results with any kind of prompts. For each case, we conduct five trials, and a detailed analysis of the responses from GPT-4 and LLaMA-3 reveals the following: GPT-4's responses are more consistent, suggesting that GPT-4 genuinely understands the prompts and examples. These instructions ``activate" GPT-4 to consider different perspectives and provide correct results. In contrast, LLaMA-3 more likely provides varied responses to the same prompt. For more obvious anomalies, such as global point anomalies and trend anomalies (see Figure~\ref{fig:trial_example_details} in Appendix), LLaMA-3 provides more stable results with correctly identified anomalies and explanations.
In summary, we observe more emergent abilities~\citep{wei2022emergent} in GPT-4, where simple instructions can activate its capability of time series anomaly detection, leading to more accurate identification and explanation of time series anomalies. Although LLaMA-3 does not exhibit these abilities to the same extent (potentially due to its smaller parameter size compared to GPT-4), it still shows some capabilities in grasping the overall shape of time series.

\subsection{LLM against Anomaly Detection Baselines}

Given the impressive performance of GPT-4 on all trial examples with different prompts, we  now evaluate how GPT-4 performs time series anomaly detection  compared to classic anomaly detection baseline methods.

\textbf{Datasets \& Evaluation Metrics}
We evaluate the performance on four common time series anomaly detection datasets~\citep{paparrizos2022tsb}: YAHOO, ECG, SVDB, and IOPS, which include anomalies in monitoring services and ECG recordings. 
In our study, we carefully curate the datasets to encompass a broad spectrum of patterns. In each dataset, we select 100 distinct time series segments with length 1,080 that demonstrate maximum variability. We utilize the initial 50\% of each time series as training data. More details can be found in Appendix~\ref{app:benchmark_dataset_settings}. 
We use F-score and Range-F~\citep{paparrizos2022tsb} to evaluate the performance. Range-F is an extension of the F-score, where a detection is considered accurate if the identified anomaly falls within the same window as the actual anomaly; in this case, we set the window size to 5.

\textbf{Baseline Methods}
In our comparison, we evaluate a range of time series anomaly detection methods. This includes traditional approaches such as Isolation Forest (IForest) \citep{liu2008isolation}, Matrix Profile (MP) \citep{yeh2016matrix}, and Autoencoder \citep{sakurada2014anomaly}. Additionally, we explore forecasting-based methods, namely LSTM \citep{malhotra2015long}, Prophet~\citep{taylor2018forecasting}, Informer~\citep{zhou2021informer}, DLinear~\citep{zeng2023transformers} and TimesNet~\citep{wu2022timesnet}. For these forecasting methods, anomalies are defined as observations deviating from the forecasted values by more than a 3-$\sigma$ (three standard deviations) threshold. More implementation details can be found in Appendix~\ref{app:baseline_settings}.

\textbf{LLM Settings}
We structure the time series segments using multi-modal prompts analogous to the example depicted in Figure~\ref{fig:example_response}, then feed the prompts to GPT-4 through the OpenAI API services\footnote{\url{https://openai.com/api/}}. Additionally, we craft a specific prompt designed to parse the output into a desired JSON format. This format encompasses two key components: a list of indices identifying the anomalous points, and a textual explanation that elucidates the rationale behind the identification of these anomalies. More details can be found in Appendix~\ref{app:llm_settings}.

\textbf{Comparison Results}
Table~\ref{tab:comp_gpt4} provides a comparative analysis of GPT-4 against baseline methods in anomaly detection tasks. Generally, we observe that most baseline methods perform well on specific datasets. For example, LSTM performs well on the YAHOO and IOPS datasets but poorly on the ECG and SVDB datasets. Conversely, the Autoencoder performs well on the ECG and SVDB datasets but poorly on YAHOO and IOPS. Compared to these baseline methods, GPT-4 shows notable achievements in F-score and Range-F metrics across the ECG, SVDB, and IOPS datasets, with an average rank of 3. Despite its performance on YAHOO, GPT-4 is the most stable model across the ECG, SVDB, and IOPS datasets, even when compared to TimesNet and Prophet. This demonstrates the potential of GPT-4 as a generalized time series anomaly detector.

\begin{table*}[t]
  \caption{Comparison between GPT-4 and Classic Time Series Anomaly Detectors with rank.}
  \label{tab:comp_gpt4}
  \scriptsize
  \begin{tabular}{c ccccccccc c}
    \toprule
    \multirow{2}{*}{Method} & \multicolumn{2}{c}{YAHOO} & \multicolumn{2}{c}{ECG} & \multicolumn{2}{c}{SVDB}  & \multicolumn{2}{c}{IOPS}  & \multirow{2}{*}{avg-Rank} \\ 
    & F-score & Range-F & F-score & Range-F & F-score & Range-F & F-score & Range-F &  \\
    \midrule
    IForest 
    & 0.0271(7) & 0.1066(8) & 0.1602(9) & 0.1780(9) & 0.0839(5) & 0.0886(9) & 0.0754(4) & 0.1125(5) &  9(7,9) \\
    MP 
    & 0.0630(4) & 0.1719(4) & 0.1654(8) & 0.1808(8) & 0.1170(4) & 0.1303(4) & 0.0064(9) & 0.0160(9) &  7(7,7) \\
    AE 
    & 0.0212(8) & 0.1078(7) & 0.3199(2) & 0.3580(2) & 0.4352(1) & 0.4626(1) & 0.0667(6) & 0.0985(7) &  4(2,4) \\
    LSTM 
    & 0.1466(1) & 0.2144(1) & 0.1737(7) & 0.2012(7) & 0.0773(9) & 0.0900(8) & 0.1062(2) & 0.1395(3) &  6(5,6) \\
    Prophet 
    & 0.0455(5) & 0.1324(5) & 0.3852(1) & 0.4556(1) & 0.1745(3) & 0.2107(3) & 0.0456(8) & 0.1324(4) & 2(2,1) \\
    Informer 
    & 0.0304(6) & 0.1161(6) & 0.1842(6) & 0.2141(6) & 0.0784(8) & 0.0916(7) & 0.0474(7) & 0.0707(8) & 8(9,8) \\
    DLinear 
    & 0.1051(3) & 0.1760(3) & 0.1867(4) & 0.2208(4) & 0.0801(7) & 0.0954(5) & 0.0676(5) & 0.0995(6) & 5(5,5) \\
    TimesNet 
    & 0.1457(2) & 0.2112(2) & 0.1867(4) & 0.2178(5) & 0.0809(6) & 0.0943(6) & 0.1443(1) & 0.1889(1) & 1(1,2) \\ \midrule
    \textbf{GPT-4} 
    & 0.0204(9) & 0.0936(9) & 0.2911(3) & 0.3258(3) & 0.2681(2) & 0.2945(2) & 0.1020(3) & 0.1414(2) & 3(2,3) \\
  \bottomrule
  \vspace{-10pt}
\end{tabular}
\end{table*}

\setlength{\columnsep}{20pt}

\begin{wraptable}{r}{0.35\textwidth} 
  \centering
  \vspace{-0.3cm}
  \caption{Details of hallucination of GPT-4 on each dataset. \textit{Mean} and \textit{Median} stand for the mean and median numbers of hallucinated points over the hallucinated segments.}
  \label{tab:hallucination_stats}
  \scriptsize
  \begin{tabular}{cccc}
    \toprule
        Dataset & \# Segments & Mean & Median \\ \midrule
        YAHOO & 24 & 122.0 & 2 \\ 
        ECG & 29 & 322.4 & 460 \\
        SVDB & 21 & 283.4 & 260 \\
        IOPS & 28 & 111.5 & 3 \\
        \bottomrule
    \end{tabular}
\end{wraptable}

\textbf{Hallucination in Indices} 
Although the F-score and Range-F results from GPT-4 appear promising, our analysis reveals that the model occasionally generates indices outside the expected time series segments, such as 1,200 for a series of length 1,000. As shown in Table~\ref{tab:hallucination_stats}, these hallucinations occur in approximately 21\% to 29\% of time series segments across different datasets. 
In datasets like YAHOO and IOPS, with a lower anomaly proportion (about 1\% to 4\% of segment points), the median number of hallucinated points is relatively low (2 or 3). However, the average number of hallucinations is significantly higher, indicating that some segments experience many spurious predictions. This issue is more pronounced in datasets such as ECG and SVDB, which have a higher anomaly ratio (about 20\% to 30\% of segment points), leading to an increase in hallucinated positions.

\begin{figure*}[t]
\centering
\small
\begin{minipage}[t]{0.32\linewidth}
\includegraphics[width=\linewidth]{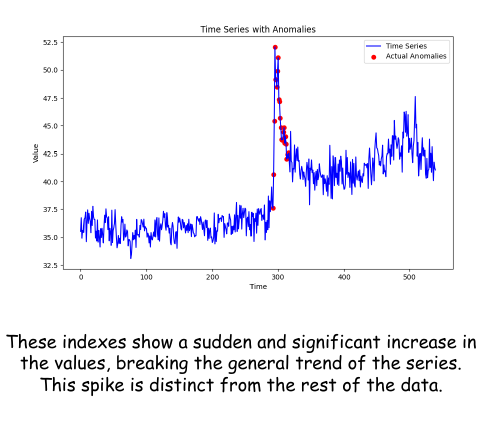}
\centering{(a) Good}
\end{minipage}
\begin{minipage}[t]{0.32\linewidth}
\includegraphics[width=\linewidth]{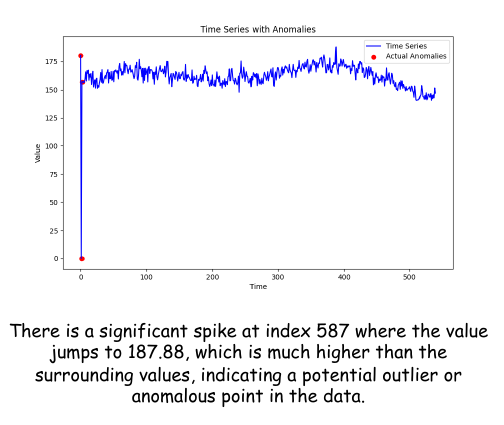}
\centering{(b) Bad}
\end{minipage}
\begin{minipage}[t]{0.32\linewidth}
\includegraphics[width=\linewidth]{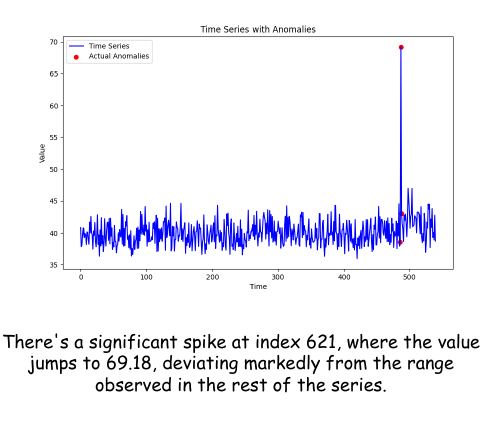}
\centering{(c) Hallucination}
\end{minipage}
\caption{Examples for a) good, b) bad, and c) hallucinated explanation by GPT-4 on IOPS dataset.}
\vspace{-5pt}
\label{fig:example_explanation_IOPS}
\end{figure*}

\textbf{Explanation Analysis} For the explanation provided by GPT-4, we manually analyze the results and classify the explanations into three categories: i) good explanation, where the model provides reasonable reasons for the detection results and correctly identifies the index of the anomalies; ii) bad explanation, where the model fails to explain the detected anomalies well or cannot detect them; and iii) good explanation with hallucination in values, where the model reasonably explains the detected anomalies but incorrectly mentions the index or value of the anomalies. Figure~\ref{fig:example_explanation_IOPS} shows examples for the three conditions, and more examples for other datasets can be found in Appendix~\ref{app:evaluation_gpt4}. Table~\ref{tab:explanation} shows the counts for explanation performance under different conditions.
For the YAHOO and IOPS datasets, which exhibit more local and global point anomalies like spikes and dips, GPT-4 generally provides accurate explanations. Poor explanation on the datasets typically occurs when GPT-4 fails to detect anomalies precisely, such as identifying only certain local point anomalies while overlooking significant global ones or missing pattern change anomalies. Occasionally, GPT-4 misinterprets figures, mistaking a dip for a spike, as shown in Figure~\ref{fig:example_explanation_IOPS}. In contrast, the ECG and SVDB datasets, which contain more context-aware anomalies such as pattern changes, pose greater challenges for GPT-4 in providing accurate explanations. The model often continues to search for local and global point anomalies, sometimes mistakenly identifying periodic spikes in ECG signals as anomalies.
\begin{wraptable}{r}{0.62\textwidth}
    \centering
    \scriptsize
    \caption{Summary of the explanation capability of GPT-4 on different datasets. }
    \begin{tabular}{ccccc}
    \toprule
        Count & YAHOO & ECG & SVDB & IOPS \\ \midrule
        Good Explanation & 25 & 26 & 9 & 25 \\ 
        Good Explanation w Hallucination & 24 & 5 & 3 & 35 \\
        \bottomrule
    \end{tabular}
    \label{tab:explanation}
\end{wraptable}
 However, when a pattern change is pronounced, GPT-4 can detect the shift and provide a coherent explanation. 
In general, hallucinations in explanation typically occur when GPT-4 describes the index or values of anomalies. Such errors are more frequent in the YAHOO and IOPS datasets, characterized by a higher incidence of point-based anomalies, but have hallucination when mentioning the indices of the anomalies or the values of the anomalies.

\textbf{Summary}
With minimal instructions, GPT-4 often presents as a great explainable time series anomaly detector, sometimes ranking in the top three among various baseline methods. However, detecting longer time series poses significant challenges to GPT-4. While it excels in identifying local and global anomalies, it struggles with more nuanced, context-dependent scenarios and tends to hallucinate about both the anomaly indices and the explanations provided.
Conversely, prompt strategies do not benefit LLaMA3, likely due to its smaller size compared to GPT-4. In the next section, we will explore whether LLaMA3's performance can be improved through instruction fine-tuning.

\section{Can LLMs be Improved via Instruction Fine-tuning?}
\label{sec:fine-tuning}
While GPT-4 can be ``activated" as an effective explainable time series anomaly detector, particularly for shorter time series, LLaMA-3 does not benefit as much from prompt engineering, primarily due to its smaller parameter size. Therefore, we aim to investigate whether LLaMA-3's performance can be improved via fine-tuning. Given the scarcity of time series with anomalies and corresponding textual explanation datasets, we propose a time series and text explanation generator TTGenerator to create the instruction datasets for fine-tuning LLaMA-3.


\subsection{Time Series and Text Explanation Generator: TTGenerator}
\label{sec:ttgenerator}

\textbf{Base Time Series Generation}
Formally, a time series dataset $X$ with $T$ timestamps can be represented as an ordered sequence of data points: $X = (x_1, x_2, \cdots, x_T)$, where $x_i$ is the data point at timestamp $i$ ($i\in T$). 
Generally, a time series is viewed as a combination of trend, seasonality, and noise components:
\begin{equation}\label{eq:decomposition}
    X = s(T) + \tau(T) + \epsilon
\end{equation}
where $s(\cdot)$ represents the base shapelet function approximating the detrended series, which could be a combination of sine and square wave functions, i.e., $\sum_n (A_n \sin (2\pi \omega_n T))$, where $A$ is the amplitude and $\omega_n$ as the frequency. Alternatively, time series can be generated via Inverse Fast Fourier Transform (IFFT), i.e., $\sum_n (A_n \exp^{\frac{2\pi \omega_n n}{N} i})$; $\tau(\cdot)$ models the overall trend of the series, which could be linear or exponential; and $\epsilon$ represents the noises which could be just white noises.

\textbf{\textbf{Anomaly Points Generation}}
Following~\cite{lai2021revisiting}, we examine various types of time series anomalies, including point-wise and pattern-wise anomalies. 
\textit{Point-wise anomalies} are defined as unexpected incidents at individual time points:
\begin{equation}\label{eq:point_anomaly}
    |x_t-\hat{x_t}| > \delta
\end{equation}
This includes local point anomalies, where $\delta = \lambda \cdot \sigma(X_{[x-C \leq x \leq x+C]})$ with $C$ as the context window size, and global point anomalies, where $\delta = \lambda \cdot \sigma(X)$, representing significant spikes or dips in the time series. Here, $\sigma$ denotes the standard deviation and $\lambda$ sets the threshold level.
\textit{Pattern-wise anomalies} represent anomalous subsequences characterized by changes in seasonality, trend, or shape. Specifically, within a time series data $X$, an underlying subsequence $X_{i,j}$ from timestamp $i$ to $j$ can be considered anomalous if:
\begin{equation}\label{eq:context_anomaly}
    \text{sim}(X_{i,j}, \hat{X}_{i,j}) > \delta
\end{equation}
This indicates significant deviation from the expected values $\hat{X}_{i,j}$. A seasonality anomaly may occur with an amplitude change (i.e., a modified $\widetilde{A}_n$ in $s(T_{i,j})$) or a period change (i.e., a modified $\widetilde{\omega}_n$ in $s(T_{i,j})$). Trend anomalies may involve a change point (where trends differ before and after point $i$, with $1<i<N$), or a trend break (where the trend changes at $i$ and then reverts at $j$, with $1<i<j<N$). Shape change anomalies may manifest as a pattern change (where the base pattern shifts starting at $i$ and continues to $j$, with $1<i<N$), or a pattern break (where the base pattern changes at $i$ but returns to normal by $j$, with $1<i<j<N$).

\textbf{\textbf{Explanation Generation}}
After generating the base time series and the anomalies, we utilize a template to produce a description of the time series. This description includes: (i) details about the base time series such as seasonality, trend, and noise; and (ii) specifics about the anomalies, including the types of anomalies and their starting and ending indices. For time series that do not contain anomalies, the description will state: \textit{``There is no obvious anomaly in this time series"}. 
To enhance the diversity of the dataset, we employ GPT-4 to rewrite the description for each sample.

In summary,  TTGenerator synthesizes time series with outliers by (i) selecting random seasonality and trend patterns, (ii) inserting various types of outliers, and (iii) generating descriptions for the time series and the anomalies. More details are provided in Appendix~\ref{app:ttgenerator}.

\subsection{Instruction Fine-tuning on LLaMA3}
With TTGenerator, we generate the instruction dataset as follows: 1) Random Selection of Length: We randomly select the length of the generated time series from various time series lengths. We do not consider longer time series due to the context window length limitation for the LLMs, such as the 8k token limit for LLaMA3.
2) Sample Generation: We generate a single sample that includes the time series values, labels for the anomalies, and explanations for both the base time series and the anomalies.
3) Text Prompt Formation: We concatenate the information of a time series to form the text prompt to train the model as: 

\begin{tcolorbox}[colback=white,colframe=white!75!black]
\tiny
\textbf{Instruction}: \{instruction\}

\textbf{Time Series Values}: \{time series values\}

\textbf{Requirements}: \{requirements\}

\textbf{Response}: \{JSON format with keys \texttt{anomaly} as the labels for the anomalies and \texttt{reason} as the explanation for the anomalies\}
\end{tcolorbox}


where \{instruction\} refers to the general instruction and \{requirements\} specify that the output should be formatted in JSON with \texttt{anomaly} and \texttt{reason} as the two keys.
4) Repetition: Finally, we repeat the procedures 1)-3) $n$ times to create the final dataset, where $n$ is the dataset size.
To fine-tune the instruction dataset on LLaMA3, we use a parameter-efficient fine-tuning (PEFT) approach, specifically LoRA~\citep{hu2021lora}, to obtain the fine-tuned model. More details on the fine-tuning can be found in Appendix~\ref{app:llm_settings}.

\subsection{Results}
We evaluate the performance on three synthesized datasets generated by TTGenerator for five types of time series anomalies: global point anomaly, local point anomaly, seasonality anomaly, trend anomaly, and shape anomaly. The datasets have different time series lengths of 100, 200, and 400. More details about the synthesized datasets can be found in Appendix~\ref{app:ttgenerator} and more experimental results in Appendix ~\ref{app:llama3_results}.

\textbf{Overall Performance}
We compare the performance of the original LLaMA3 with our fine-tuned version. The results are shown in Table~\ref{tab:llama3_overall_performance}. Generally, for both models, the performance decreases as the length of the time series increases. Despite the relatively low F-score for point-aware anomalies, the Range-F score is relatively high, indicating that the model is able to capture the correct anomalies but may hallucinate in the surrounding indices.
Comparing point-aware and context-aware anomalies, both models provide more stable performance on context-aware anomalies compared to point-aware anomalies across different time series lengths.
Comparing the original and fine-tuned versions of LLaMA3, the ability to detect local point anomalies and shape anomalies does not seem to benefit much from the instruction fine-tuning. However, we observe general improvements in the average F-score and Range-F, with significant improvements in detecting seasonality anomalies.

\begin{figure*}[t]
\centering
\small
\begin{minipage}[t]{0.3\linewidth}
\includegraphics[width=\linewidth]{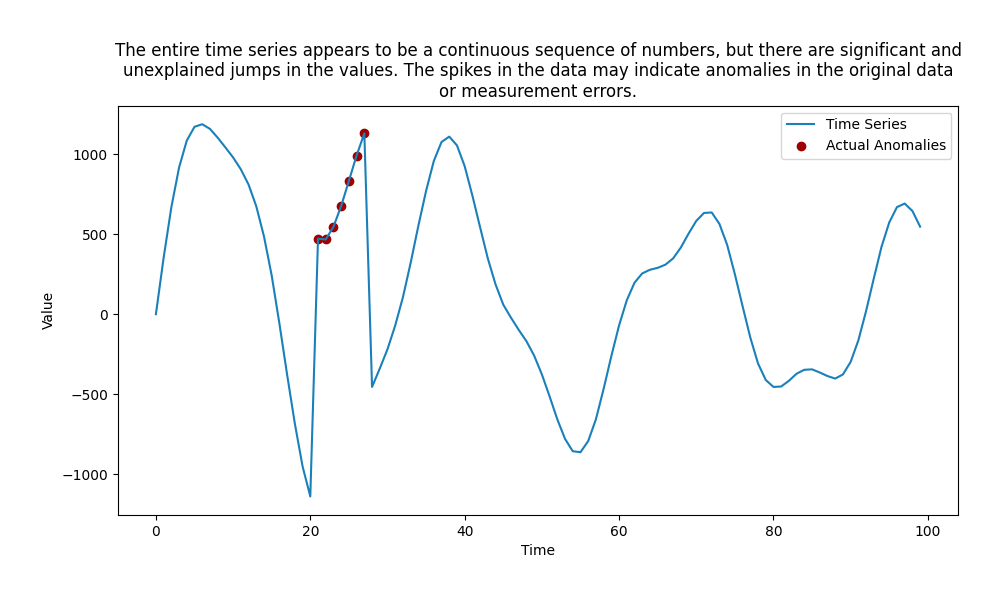}
\centering{(a) Good}
\end{minipage}
\begin{minipage}[t]{0.3\linewidth}
\includegraphics[width=\linewidth]{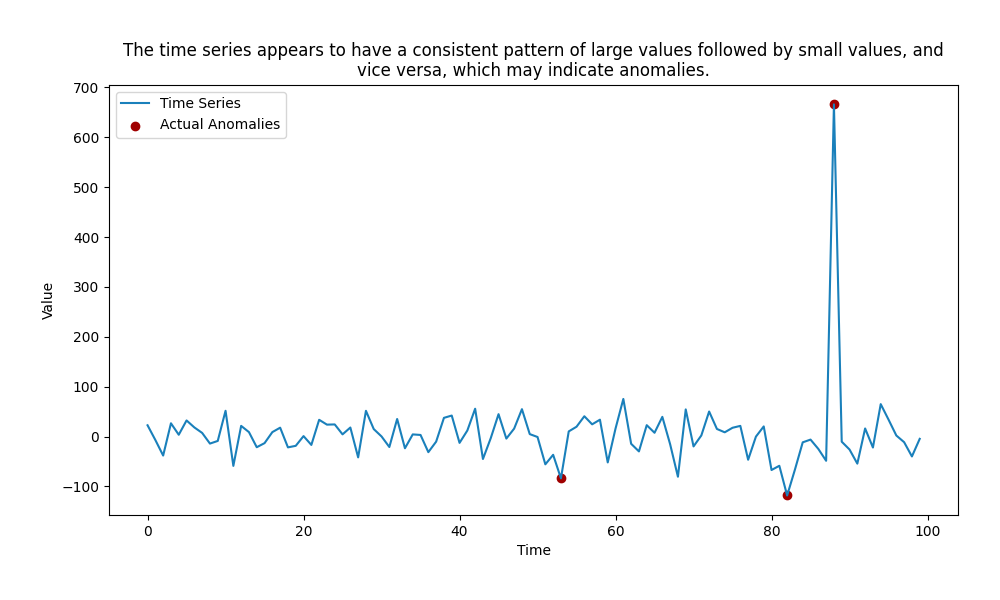}
\centering{(b) Bad}
\end{minipage}
\begin{minipage}[t]{0.3\linewidth}
\includegraphics[width=\linewidth]{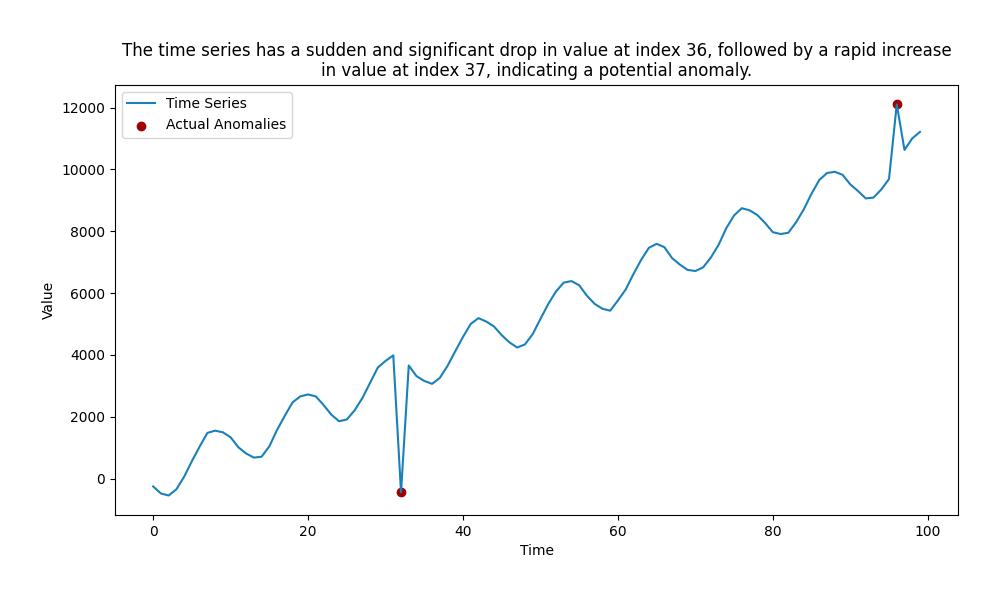}
\centering{(c) Hallucination}
\end{minipage}
\caption{Examples for a) good, b) bad, and c) hallucinated explanation by LLaMA3}
\label{fig:example_explanation_llama3}
\end{figure*}

\begin{table}[t] 
  \centering
    \centering
    \tiny
    \caption{Overall performance of original and fine-tuned LLaMA3 on the synthesized dataset}
    \begin{tabular}{cccccccc}
    \toprule
    \multirow{2}{*}{Anomaly Type} & \multirow{2}{*}{Metrics} &  \multicolumn{3}{c}{Original} & \multicolumn{3}{c}{Fine-tuned} \\ 
    & & 100 & 200 & 400 & 100 & 200 & 400 \\
    \midrule
    \multirow{2}{*}{Global Point} 
    & F-score & 0.0528 & 0.0300 & 0.0104 & 0.0539 & 0.0264 & 0.0130 \\
    & Range-F & 0.4091 & 0.2205 & 0.1207 & 0.4157 & 0.2443 & 0.1421 \\ \midrule
    \multirow{2}{*}{Local Point} 
    & F-score & 0.0415 & 0.0286 & 0.0182 & 0.0426 & 0.0245 & 0.0174 \\ 
    & Range-F & 0.3481 & 0.2595 & 0.1664 & 0.3611 & 0.2424 & 0.1624 \\ \midrule
    \multirow{2}{*}{Seasonality} 
    & F-score & 0.2706 & 0.2890 & 0.2119 & 0.4189 & 0.3448 & 0.2121 \\ 
    & Range-F & 0.4436 & 0.3904 & 0.3369 & 0.5479 & 0.4114 & 0.3477 \\ \midrule
    \multirow{2}{*}{Trend} 
    & F-score & 0.0772 & 0.1315 & 0.0958 & 0.0750 & 0.1314 & 0.1871 \\ 
    & Range-F & 0.2520 & 0.2195 & 0.1367 & 0.2321 & 0.2397 & 0.2354 \\ \midrule
    \multirow{2}{*}{Shape} 
    & F-score & 0.2193 & 0.2171 & 0.2460 & 0.2489 & 0.1966 & 0.2414 \\ 
    & Range-F & 0.3981 & 0.3199 & 0.3115 & 0.3871 & 0.3117 & 0.3393 \\ \midrule
    \multirow{2}{*}{Overall} 
    & F-score & 0.1323 & 0.1392 & 0.1165 & 0.1679 & 0.1447 & 0.1342 \\ 
    & Range-F & 0.3702 & 0.2820 & 0.2144 & 0.3888 & 0.2899 & 0.2454 \\ 
    \bottomrule
    \end{tabular}
    \label{tab:llama3_overall_performance}
\end{table}

\textbf{Hallucination in Indices} Similar to GPT-4, LLaMA3 exhibits hallucinations in its detection results. Table~\ref{tab:hallucination_stats_llama3} presents these findings. Interestingly, we observe a decreasing trend in the number of hallucinated time series segments with the fine-tuned LLaMA3 compared to GPT-4's performance (Appendix~\ref{app:evaluation_gpt4}), while LLaMA3 tends to hallucinate more indices than GPT-4.

\begin{table}[h]
    \centering
    \begin{minipage}{0.38\linewidth}
        \centering
\caption{Details of hallucination of fine-tuned LLaMA3 on each dataset. }
  \label{tab:hallucination_stats_llama3}
  \tiny
  \begin{tabular}{ccccc}
    \toprule
         Length & \# Segments & Mean & Median \\ \midrule
         100 & 63 & 197.5 & 166.0 \\
         200 & 43 & 213.2 & 203.5 \\
         400 & 36 & 148.9 & 141.0 \\
        \bottomrule
    \end{tabular}
    \end{minipage}\hfill
    \begin{minipage}{0.6\linewidth}
        \centering
 \caption{Summary of the explanation capability of LLaMA3 for different anomaly types. }
 \tiny
 \label{tab:explanation_llama3}
    \begin{tabular}{cccccc}
    \toprule
        Count & Global & Local & Seasonal & Trend & Shape \\ \midrule
        Good Explanation & 27 & 23 & 7 & 9 & 12 \\ 
        Good Explanation w Hallucination & 3 & 5 & 3 & 3 & 1  \\
        \bottomrule
    \end{tabular}
    \end{minipage}
\end{table}

\textbf{Explanation Analysis}
We further manually check the explanation performance by the fine-tuned LLaMA3 on the five anomaly types over the three datasets, where there are 60 samples for each anomaly type. The results are shown in Table~\ref{tab:explanation_llama3}. Similar to GPT-4, we notice that the model explains well for point-aware anomalies than the context-aware anomalies. Different from GPT-4, the explanations provided by LLaMA3 are more general, such as "\textit{The anomalies are mostly due to the sudden changes in the value of the time series}". Examples for the explanations provided by LLaMA3 can be found in Figure~\ref{fig:example_explanation_llama3}. 

\section{Conclusion}
In this paper, we comprehensively investigate the capability of Large Language Models (LLMs) in time series anomaly detection by addressing three key questions: Can LLMs be directly applied for explainable time series anomaly detection? How can LLMs detect and explain time series anomalies via prompt engineering? Can we improve LLMs' detection performance through instruction fine-tuning? The answers to these questions are: No, Yes, and Yes, respectively, with evidence showing that GPT-4 demonstrates competent performance compared to baseline methods with minimal effort in prompt engineering, and LLaMA3 achieves better performance after instruction fine-tuning. In summary, LLMs show promising potential for time series anomaly detection, while customized prompts and instructions are essential.


\bibliographystyle{plainnat}
\bibliography{main}

\newpage
\appendix

\section{Experimental Settings}

\subsection{Benchmark Dataset Settings}
\label{app:benchmark_dataset_settings}
We selected four widely used time series anomaly detection datasets: YAHOO, ECG, SVDB, and IOPS, as referenced in the paper by Paparrizos et al. (2022)\citep{paparrizos2022tsb}. The original datasets can be downloaded from the repository\footnote{\url{https://github.com/TheDatumOrg/TSB-UAD}}.
We constructed the evaluation dataset by manually selecting segments from the time series data. First, we determined the window size for each time series using the Fast Fourier Transform (FFT) and then computed the median window size across the dataset. The segment length was set to four times the median window size, resulting in a segment length of 1080 based on the window sizes of the four datasets. Each time series was partitioned into multiple segments of this length.
We manually inspected the segments with a length of 1080 for each dataset, selecting time series with diverse distributions. From these, we randomly extracted 100 segments for evaluation. Table~\ref{tab:dataset_details} presents the specifics of segment lengths and the number of time series used, where '\# Time Series' denotes the distinct time series in the original dataset, and '\% Anomalies' denotes the average proportion of anomalies in the selected segments. Examples for each dataset can be seen in Figure~\ref{fig:example_explanation_IOPS} and Figure~\ref{fig:example_explanation_all}.
\begin{table}[h]
    \centering
    \small
    \caption{Dataset Details}
    \begin{tabular}{cccc}
    \toprule
        Dataset & Segment Length & \# Time Series & \% Anomalies\\ \midrule
        YAHOO & 1080 & 100 & 1.42 \\ 
        ECG & 1080 & 21 & 19.99 \\
        SVDB & 1080 & 21 & 27.94 \\
        IOPS & 1080 & 21 & 3.78 \\ \bottomrule
    \end{tabular}
    \label{tab:dataset_details}
\end{table}

\subsection{Baselines Settings}
\label{app:baseline_settings}

\textbf{IForest\citep{liu2008isolation}}
We use the Scikit-learn implementation\footnote{\url{https://scikit-learn.org/stable/}} with \texttt{n\_estimators} set to 100. Following the approach in \cite{wu2022timesnet}, we employ the Fast Fourier Transform to determine the optimal window size for each time series.

\textbf{Matrix Profile (MP)~\citep{yeh2016matrix}}
We use the Stumpy implementation\footnote{\url{https://stumpy.readthedocs.io/en/latest/index.html}} and set the window size for each time series based on the Fast Fourier Transform strategy.

\textbf{Autoencoder (AE)~\citep{sakurada2014anomaly}}
Following the parameter settings suggested in \cite{paparrizos2022tsb}, we use three encoder and three decoder layers with ReLU as the activation function. The window size is adjusted to match the length of the test data.

\textbf{Prophet~\citep{taylor2018forecasting}}
We use the official Facebook implementation\footnote{\url{https://facebook.github.io/prophet/}} and detect anomalies using the forecasted \texttt{yhat\_upper} and \texttt{yhat\_lower} bounds.

\paragraph{LSTM~\citep{malhotra2015long}, Informer~\citep{zhou2021informer}, TimesNet~\citep{wu2022timesnet}, and DLinear~\citep{zeng2023transformers}}
Implementations are sourced from NeuralForecast\footnote{\url{https://nixtlaverse.nixtla.io/}}. Anomalies are detected by applying the 3-$\sigma$ rule, which flags any data point deviating more than three standard deviations from the mean.

\begin{figure}[t]
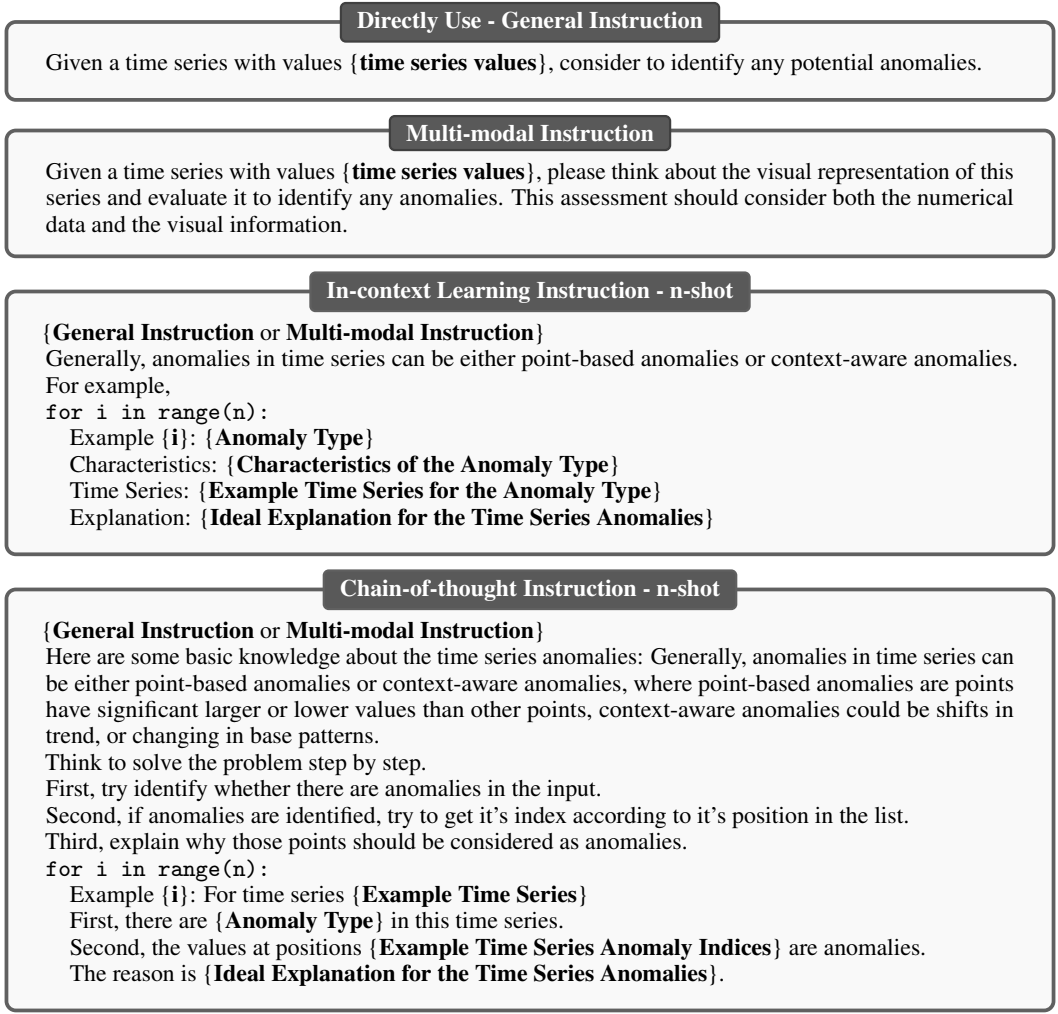

    \centering
    \small
    \begin{tcolorbox}[enhanced,attach boxed title to top center={yshift=-3mm,yshifttext=-1mm},
  colback=gray!5!white,colframe=gray!75!black,colbacktitle=gray!70!black,
  title=Directly Use - General Instruction,fonttitle=\bfseries,
  boxed title style={size=small,colframe=gray!50!black} ]
  Given a time series with values \{\textbf{time series values}\}, consider to identify any potential anomalies. 
\end{tcolorbox}
    \begin{tcolorbox}[enhanced,attach boxed title to top center={yshift=-3mm,yshifttext=-1mm},
  colback=gray!5!white,colframe=gray!75!black,colbacktitle=gray!70!black,
  title=Multi-modal Instruction,fonttitle=\bfseries,
  boxed title style={size=small,colframe=gray!50!black} ]
  Given a time series with values \{\textbf{time series values}\}, please think about the visual representation of this series and evaluate it to identify any anomalies. This assessment should consider both the numerical data and the visual information.
\end{tcolorbox}
\begin{tcolorbox}[enhanced,attach boxed title to top center={yshift=-3mm,yshifttext=-1mm},
  colback=gray!5!white,colframe=gray!75!black,colbacktitle=gray!70!black,
  title=In-context Learning Instruction - n-shot,fonttitle=\bfseries,
  boxed title style={size=small,colframe=gray!75!black} ]
\{\textbf{General Instruction} or \textbf{Multi-modal Instruction}\} 
  
Generally, anomalies in time series can be either point-based anomalies or context-aware anomalies. For example, 

\texttt{for i in range(n):}

\quad Example \{\textbf{i}\}: \{\textbf{Anomaly Type}\}

\quad Characteristics: \{\textbf{Characteristics of the Anomaly Type}\}

\quad Time Series: \{\textbf{Example Time Series for the Anomaly Type}\}

\quad Explanation: \{\textbf{Ideal Explanation for the Time Series Anomalies}\}

\end{tcolorbox}
\begin{tcolorbox}[enhanced,attach boxed title to top center={yshift=-3mm,yshifttext=-1mm},
  colback=gray!5!white,colframe=gray!75!black,colbacktitle=gray!70!black,
  title=Chain-of-thought Instruction - n-shot,fonttitle=\bfseries,
  boxed title style={size=small,colframe=gray!75!black} ]
\{\textbf{General Instruction} or \textbf{Multi-modal Instruction}\} 
  
Here are some basic knowledge about the time series anomalies: 
Generally, anomalies in time series can be either point-based anomalies or context-aware anomalies, where point-based anomalies are points have significant larger or lower values than other points, context-aware anomalies could be shifts in trend, or changing in base patterns.

Think to solve the problem step by step. 

First, try identify whether there are anomalies in the input. 

Second, if anomalies are identified, try to get it's index according to it's position in the list. 

Third, explain why those points should be considered as anomalies. 

\texttt{for i in range(n):}

\quad Example \{\textbf{i}\}: For time series \{\textbf{Example Time Series}\}

\quad First, there are \{\textbf{Anomaly Type}\} in this time series.

\quad Second, the values at positions \{\textbf{Example Time Series Anomaly Indices}\} are anomalies. 

\quad The reason is \{\textbf{Ideal Explanation for the Time Series Anomalies}\}. 
    
\end{tcolorbox}
    \caption{Full Instruction Prompt for Each Strategy. For in-context learning and chain-of-thought learning, either the general instruction or multi-modal instruction is added to the beginning of the prompt to guide LLMs in performing the anomaly detection task.
}
    \label{fig:instruction_details}
\end{figure}

\begin{figure}[t]
    \centering
    \small
    \begin{tcolorbox}[enhanced,attach boxed title to top center={yshift=-3mm,yshifttext=-1mm},
  colback=gray!5!white,colframe=gray!75!black,colbacktitle=gray!70!black,
  title=Requirements for Trial Cases,fonttitle=\bfseries,
  boxed title style={size=small,colframe=gray!50!black} ]
If anomalies are present, please indicate: 1) The presence of anomaly points in this time series. 2) The indices of these anomaly points, and 3) The reasoning behind these points being considered anomalies. 
\end{tcolorbox}
    \begin{tcolorbox}[enhanced,attach boxed title to top center={yshift=-3mm,yshifttext=-1mm},
  colback=gray!5!white,colframe=gray!75!black,colbacktitle=gray!70!black,
  title=Requirements for General Experiments,fonttitle=\bfseries,
  boxed title style={size=small,colframe=gray!50!black} ]
Please consider answering the following questions according to your observation. 
First, please try to identify the potential anomalies, and provide the list of the indexes of anomalies, if no anomalies, please return [].
Second, if there are anomalies in the time series, please provide a short explanation of the anomalies.

Summarize the answers into two keys:

- \textbf{anomaly}: a list of indexes

- \textbf{reason}: a string of explanation

And format the output as JSON with the two keys.

Required: return the JSON only without other information.
\end{tcolorbox}
\caption{Details for the requirements for trial cases and general experiments. 
}
    \label{fig:requirements_details}
\end{figure}

\subsection{Prompt Settings}
\label{app:prompt_settings}

The prompt we used for inference contains two parts: the instruction part and the requirements part. 

\textbf{Instruction Prompt} Figure~\ref{fig:instruction_details} provides the full details of the instructions for each type of prompt strategy. The example time series used to generate the in-context learning and chain-of-thought prompts are shown in Figure~\ref{fig:trial_example_details}. Note that the range of n for in-context learning is 1-5, and for chain-of-thought learning is 0-5.

\textbf{Requirements Prompt}
The requirements are specified for either trial cases or general experiments, with full details shown in Figure~\ref{fig:requirements_details}. For general experiments, we request the LLMs to return results in JSON format to 1) facilitate easier extraction of detection results and 2) avoid generating lengthy responses that may exceed the context window length. After obtaining the JSON output, we use LangChain's~\footnote{\url{https://www.langchain.com/}} JSON output parser for further analysis.

\subsection{Trial Examples}
\label{app:trial_examples}

\begin{figure}[t]
    \centering
    \includegraphics[width=\linewidth]{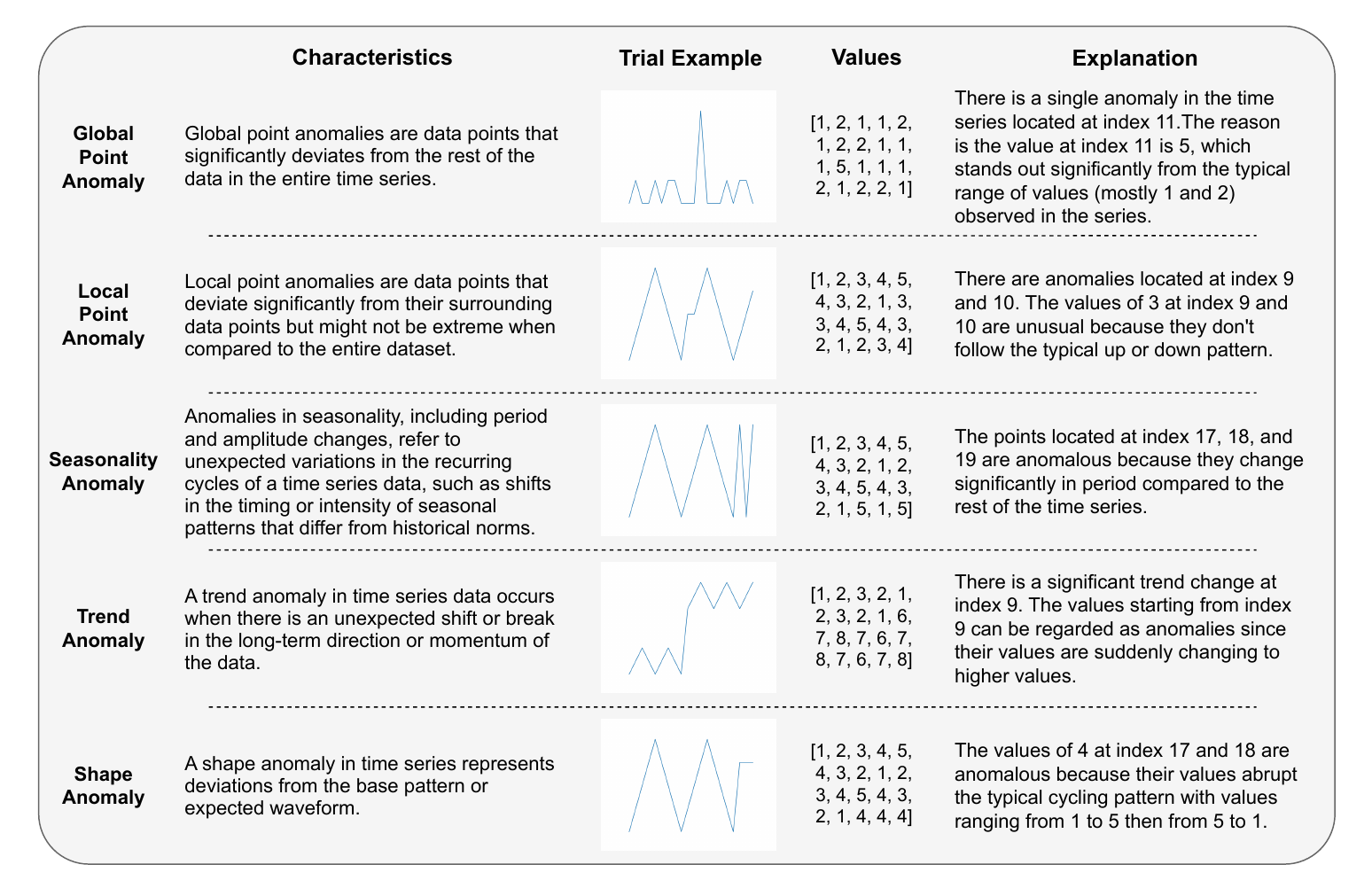}
    \caption{Details for the trial examples used for the case studies for the time series. }
    \label{fig:trial_example_details}
\end{figure}

The details about the trial examples used in Sections~\ref{sec:empirical_study} and \ref{sec:prompt_evaluation} are shown in Figure~\ref{fig:trial_example_details}, where the explanation part describes the ideal explanation for those anomalies. When constructing in-context learning and chain-of-thought prompts with n-shot examples, we use distinct anomaly types to formulate the prompt. For example, when inferring on a time series with shape anomalies, we will randomly choose examples of other types of anomalies, such as local point anomalies. Specifically, we set n to 1 to obtain the results shown in Figure~\ref{fig:example_response}.

\subsection{TTGenerator Details}
\label{app:ttgenerator}

\textbf{Base Time Series Generation}
Generally, a time series is viewed as a combination of trend, seasonality, and noise, as described in equation~\ref{eq:decomposition}. 
For the seasonality component, we use one of three methods:
i) A single sine wave function, i.e., \( A\sin (2\pi \omega T+\beta) \), where \( A \) is the amplitude (ranging from 1 to 1000), \( \omega \) is the frequency (ranging from 1 to 10), and \( \beta \) is a phase shift (ranging from 0 to \( 2\pi \)).
ii) A combination of sine wave functions, i.e., \( \sum_n (A_n \sin (2\pi \omega_n T)) \), where \( A_n = \frac{1}{2n+1} \), following the settings in \cite{lai2021revisiting}, and \( n \) is randomly sampled in the range of 3 to 10.
iii) An IFFT function, i.e., \( \sum_n (A_n \exp^{\frac{2\pi \omega_n n}{N} i}) \), where \( n \) is randomly selected in the range of 0 to 10.
To determine the seasonality for a time series, we randomly sample from these three methods with probabilities [0.25, 0.25, 0.5].
For the trend component, we consider either a linear trend, polynomial trend, or no trend, with sampling probabilities [0.3, 0.1, 0.6], assuming most time series have no trend and more linear trends than polynomial trends. For the linear trend, we randomly sample the slope in the range (-1, 1). For the polynomial trend, we randomly sample the degree in the range 2 to 5, and the coefficients for each degree are sampled from the range (-1, 1), with a shift randomly sampled from the range (-5, 5).
For the noise component, we use normally distributed noise with a mean of 0 and a standard deviation of 1.
We then combine these three components to generate the final base time series, where the amplitude for the trend is in the range (1, 200), and for the noise is in the range (1, 50).

\begin{figure}[t]
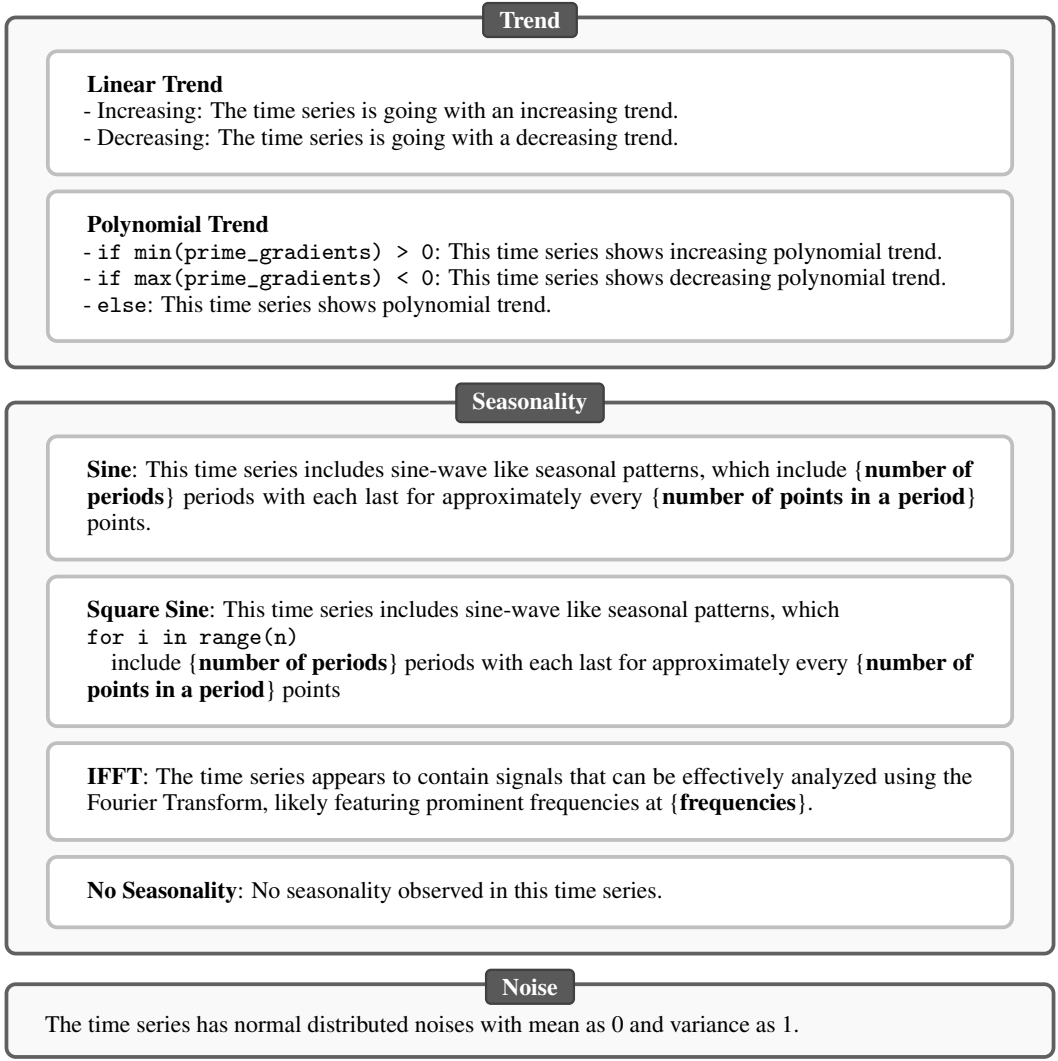

    \centering
    \small
\begin{tcolorbox}[enhanced,attach boxed title to top center={yshift=-3mm,yshifttext=-1mm},
  colback=gray!5!white,colframe=gray!75!black,colbacktitle=gray!70!black,
  title=Trend,fonttitle=\bfseries,
  boxed title style={size=small,colframe=gray!50!black} ]

\begin{tcolorbox}[colback=white,colframe=white!75!black]
\textbf{Linear Trend}

- Increasing: The time series is going with an increasing trend.

- Decreasing: The time series is going with a decreasing trend.
\end{tcolorbox}

\begin{tcolorbox}[colback=white,colframe=white!75!black]
\textbf{Polynomial Trend}

- \texttt{if min(prime\_gradients) > 0}: This time series shows increasing polynomial trend.

- \texttt{if max(prime\_gradients) < 0}: This time series shows decreasing polynomial trend.

- \texttt{else}: This time series shows polynomial trend.

\end{tcolorbox}

\end{tcolorbox}
    \begin{tcolorbox}[enhanced,attach boxed title to top center={yshift=-3mm,yshifttext=-1mm},
  colback=gray!5!white,colframe=gray!75!black,colbacktitle=gray!70!black,
  title=Seasonality,fonttitle=\bfseries,
  boxed title style={size=small,colframe=gray!50!black} ]

\begin{tcolorbox}[colback=white,colframe=white!75!black]
\textbf{Sine}: This time series includes sine-wave like seasonal patterns, which include \{\textbf{number of periods}\} periods with each last for approximately every \{\textbf{number of points in a period}\} points. 
\end{tcolorbox}

\begin{tcolorbox}[colback=white,colframe=white!75!black]
\textbf{Square Sine}: This time series includes sine-wave like seasonal patterns, which 

\texttt{for i in range(n)}

\quad include \{\textbf{number of periods}\} periods with each last for approximately every \{\textbf{number of points in a period}\} points
\end{tcolorbox}

\begin{tcolorbox}[colback=white,colframe=white!75!black]
\textbf{IFFT}: The time series appears to contain signals that can be effectively analyzed using the Fourier Transform, likely featuring prominent frequencies at \{\textbf{frequencies}\}. 
\end{tcolorbox}

\begin{tcolorbox}[colback=white,colframe=white!75!black]
\textbf{No Seasonality}: No seasonality observed in this time series.
\end{tcolorbox}

\end{tcolorbox}
    \begin{tcolorbox}[enhanced,attach boxed title to top center={yshift=-3mm,yshifttext=-1mm},
  colback=gray!5!white,colframe=gray!75!black,colbacktitle=gray!70!black,
  title=Noise,fonttitle=\bfseries,
  boxed title style={size=small,colframe=gray!50!black} ]
The time series has normal distributed noises with mean as 0 and variance as 1. 
\end{tcolorbox}
\caption{Template for time series base pattern explanation. 
}
    \label{fig:ttgenerator_base_explanation_template}
\end{figure}

\begin{figure}[t]
    \centering
    \small
    \begin{tcolorbox}[enhanced,attach boxed title to top center={yshift=-3mm,yshifttext=-1mm},
  colback=gray!5!white,colframe=gray!75!black,colbacktitle=gray!70!black,
  title=Point-aware Anomalies,fonttitle=\bfseries,
  boxed title style={size=small,colframe=gray!50!black} ]

\begin{tcolorbox}[colback=white,colframe=white!75!black,title=Global Point Anomaly,coltitle=black]
There are some point-based global anomalies in the time series, the positions are \{\textbf{position list}\}

- \texttt{if all spikes}: with significant spikes compared to the rest of the time series.

- \texttt{if all dips}: with significant dips compared to the rest of the time series. 

- \texttt{else:} with significant spikes and dips, where there are spikes in positions \{\textbf{spike positions}\} and dips in positions \{\textbf{dip positions}\}
\end{tcolorbox}

\begin{tcolorbox}[colback=white,colframe=white!75!black, title=Local Point Anomaly, coltitle=black]
There are some point-based local anomalies in the time series, with significant outlier values compared to their surrounding values, the positions are \{\textbf{position list}\}

- \texttt{if all spikes}: with significant spikes.

- \texttt{if all dips}: with significant dips.

- \texttt{else:} with significant spikes and dips, where there are spikes in positions \{\textbf{spike positions}\} and dips in positions \{\textbf{dip positions}\}
\end{tcolorbox}
\end{tcolorbox}

\caption{Template for time series anomaly pattern explanation - point-aware anomalies. 
}
    \label{fig:ttgenerator_anomaly_explanation_template_point}
\end{figure}

\begin{figure}[t]
\centering
\small
\begin{tcolorbox}[enhanced,attach boxed title to top center={yshift=-3mm,yshifttext=-1mm},
  colback=gray!5!white,colframe=gray!75!black,colbacktitle=gray!70!black,
  title=Context-aware Anomalies,fonttitle=\bfseries,
  boxed title style={size=small,colframe=gray!50!black} ]

\begin{tcolorbox}[colback=white,colframe=white!75!black,title=Seasonality Anomaly,coltitle=black]
\textbf{Amplitude Change}

- Larger Amplitude: We can observe the amplitude of the time series changes to larger values between indexes \{\textbf{start index}\} to \{\textbf{end index}\}, where the values change to about \{\textbf{ratio}\} times about the original values.

- Smaller Amplitude: We can observe the amplitude of the time series changes to smaller values between indexes \{\textbf{start index}\} to \{\textbf{end index}\}, where the values change to about \{\textbf{ratio}\} of the original values.

\tcblower

\textbf{Period Change}

- Longer Period: We can observe the seasonality period change between indexes \{\textbf{start index}\} and \{\textbf{end index}\}, where the period changes to a longer period.

- Shorter Period: We can observe the seasonality period change between indexes \{\textbf{start index}\} and \{\textbf{end index}\}, where the period changes to a shorter period.

\end{tcolorbox}

\begin{tcolorbox}[colback=white,colframe=white!75!black,title=Trend Anomaly,coltitle=black]
\textbf{Trend Change}

- If increase: We can observe a change point at index \{\textbf{change point}\} where the value increases by \{\textbf{changed value}\}.

- If decrease: We can observe a change point at index \{\textbf{change point}\} where the value decreases by \{\textbf{changed value}\}.

\tcblower
\textbf{Trend Break}

- If increase: There is a significant value increase since index \{\textbf{start index}\} and the values drop back to the original trend since index \{\textbf{end index}\}. 

- If decrease: There is a significant value decrease since index \{\textbf{start index}\} and the values increase back to the original trend since index \{\textbf{end index}\}. 

\end{tcolorbox}

\begin{tcolorbox}[colback=white,colframe=white!75!black,title=Shape Anomaly,coltitle=black]
\textbf{Shape Change}: 
There shows the base pattern of the time series change since the index \{\textbf{start index}\}, where the time series changed to \{\textbf{changed time series base pattern description}\}.
\tcblower
\textbf{Shape Break}: There are base patterns changes between the index \{\textbf{start index}\} and the \{\textbf{end index}\}, where during that time, we can observe the time series as \{\textbf{changed time series base pattern description}\}. 
\end{tcolorbox}

\end{tcolorbox}
\caption{Template for time series anomaly pattern explanation - context-aware anomalies. 
}
    \label{fig:ttgenerator_anomaly_explanation_template_context}
\end{figure}

\textbf{Anomaly Points Generation} The anomalies in time series can be roughly classified as point-aware anomalies and context-aware anomalies. 

The point-aware anomalies can be either local anomalies, where $\delta = \lambda \cdot \sigma(X_{[x-C \leq x \leq x+C]})$ with $C$ as the context window size, or global anomalies, where $\delta = \lambda \cdot \sigma(X)$. 
For global point anomalies, we set the anomaly value to be $\lambda \cdot \sigma(X)$ with $\lambda$ in the range (3, 20). For local point anomalies, we set the anomaly value to be $\lambda \cdot \sigma(X_{[x-C \leq x \leq x+C]})$ with context window $C$ in the range (10, 50) and $\lambda$ in the range (2, 5). 
We randomly sample 1-6 points in a time series to be replaced by point-wise anomaly values.

The pattern-aware anomalies can be classified into seasonality anomalies, trend anomalies, and shape anomalies. 
Specifically, a seasonality anomaly may occur with an amplitude change, i.e., a modified $\widetilde{A}_n$ in $s(T_{i,j})$. We randomly sample whether the amplitude becomes larger or smaller, setting the larger ratio to be 1.5 to 3 times the original values, and the smaller ratio to be 0.25 to 0.75 of the original values. A period change, i.e., a modified $\widetilde{\omega}_n$ in $s(T_{i,j})$, can also occur. We randomly sample whether the period becomes longer or shorter, with the longer ratio being 1.5 to 3, and the shorter ratio being 0.25 to 0.75. 

Trend anomalies occur when there is a change point where trends differ before and after point $i$, with $1 < i < N$. We set the change point between 0.2 to 0.8 of the time series and randomly select whether the trend is increasing or decreasing by 1.5 to 5 times the standard deviation of the base time series. A trend break occurs where the trend changes at $i$ and then reverts at $j$, with $1 < i < j < N$. We set $i$ in the 0.2 to 0.8 position of the time series and $j-i$ to be about 0.05 to 0.2 of the length of the time series. We randomly decide whether the trend break is increasing or decreasing, with the change being 1.5 to 5 times the standard deviation of the base time series.

Shape anomalies may occur as a pattern change, where the base pattern shifts starting at $i$, with $1 < i < N$. We randomly select the start point in the 0.2 to 0.6 range of the original time series. A pattern break occurs where the base pattern changes at $i$ but returns to normal by $j$, with $1 < i < j < N$. We set the start point $i$ in the 0.2 to 0.6 range of the original time series, and the break length to be 0.2 to 0.4 of the original time series. Specifically, to generate shape anomalies, we use a different seasonality type from the original time series and reuse TTGenerator to create a new base time series with the targeted seasonality type. For example, if the original time series has sine wave seasonality, we generate a new time series with IFFT seasonality to insert as shape anomalies. 

To generate anomalies in a time series, we first randomly select 1-3 types from the five anomaly categories: global point anomaly, local point anomaly, seasonality anomaly, shape anomaly, and trend anomaly. If a seasonality, shape, or trend anomaly is selected, we further specify the downstream anomaly type. For example, for a trend anomaly, we randomly select whether it is a shape change or a shape break.

\textbf{Explanation Generation}
The templates used to generate descriptions for the base time series are shown in Figure~\ref{fig:ttgenerator_base_explanation_template}. We concatenate the descriptions for the trend, seasonality, and noise components to form the overall description of the base time series. Note that if the time series lacks a trend, we omit any trend-related description. 
The templates used to generate explanations for the anomalies are shown in Figure~\ref{fig:ttgenerator_anomaly_explanation_template_point} and Figure~\ref{fig:ttgenerator_anomaly_explanation_template_context}. We concatenate the descriptions of the anomalies in a time series to form the overall anomaly explanation for that time series.

\begin{figure}
    \centering
\begin{minipage}[t]{0.49\linewidth}
\includegraphics[width=\linewidth]{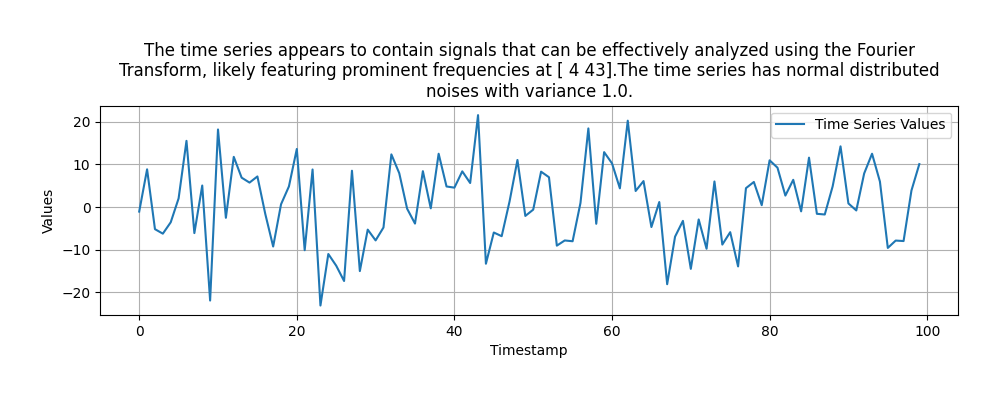}
\centering{(a)}
\end{minipage}
\begin{minipage}[t]{0.49\linewidth}
\includegraphics[width=\linewidth]{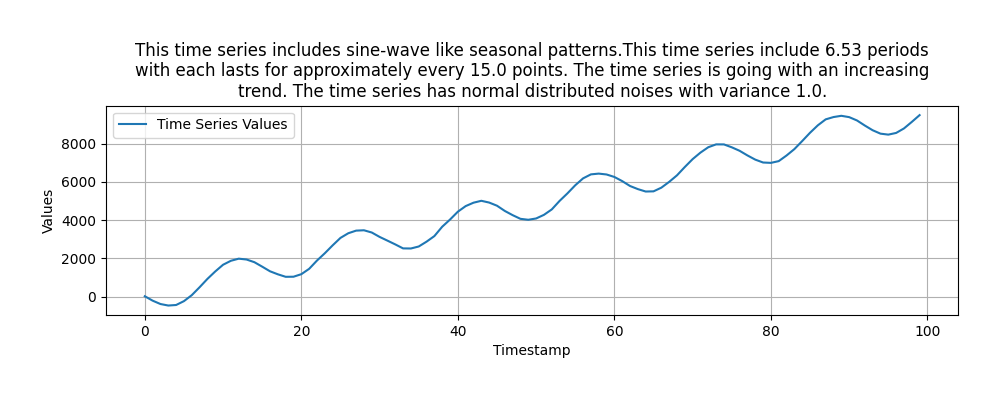}
\centering{(b)}
\end{minipage}
    \caption{Examples for the base time series with automatically generated explanation. (a) Example for a time series with seasonality as IFFT and without trend. (b) Example for a time series with sine-wave like seasonality and linear increasing trend.}
    \label{fig:example_base_generator}
\end{figure}

\begin{figure}
    \centering
\begin{minipage}[t]{0.49\linewidth}
\includegraphics[width=\linewidth]{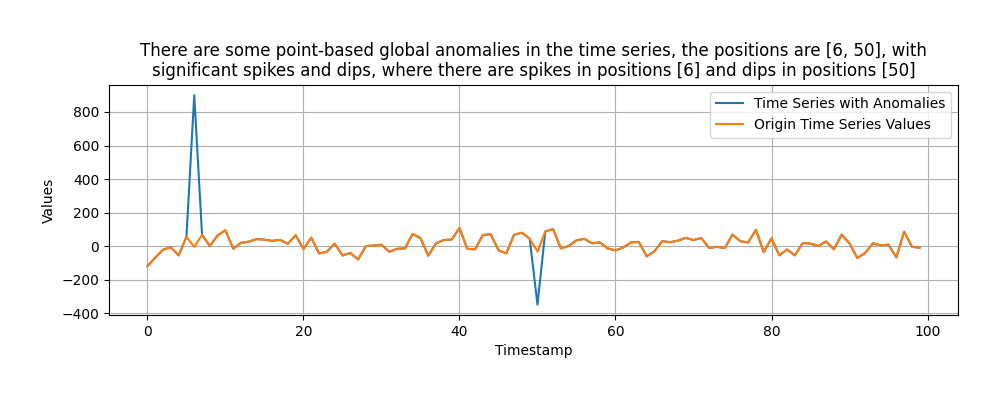}
\centering{(a) Global point anomaly}
\end{minipage}
\begin{minipage}[t]{0.49\linewidth}
\includegraphics[width=\linewidth]{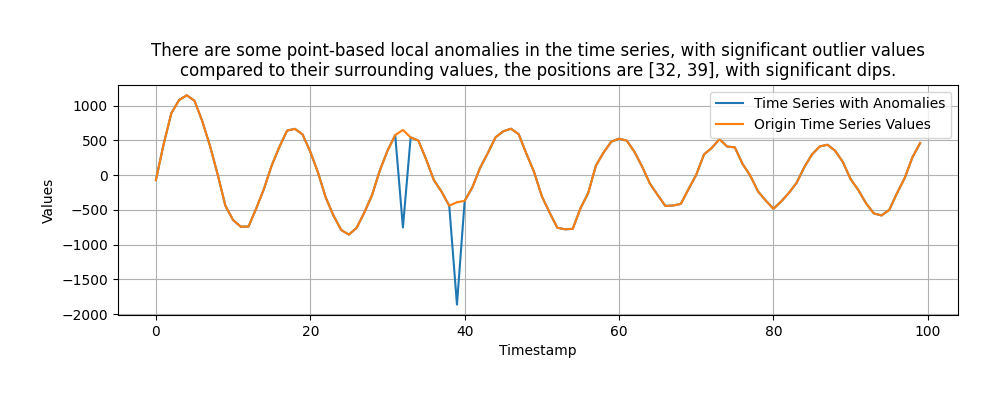}
\centering{(b) Local point anomaly}
\end{minipage}
    \caption{Examples for the time series with point anomalies with automatically generated anomaly explanation.}
    \label{fig:example_anomaly_point}
\end{figure}

\begin{figure}
    \centering
\begin{minipage}[t]{0.49\linewidth}
\includegraphics[width=\linewidth]{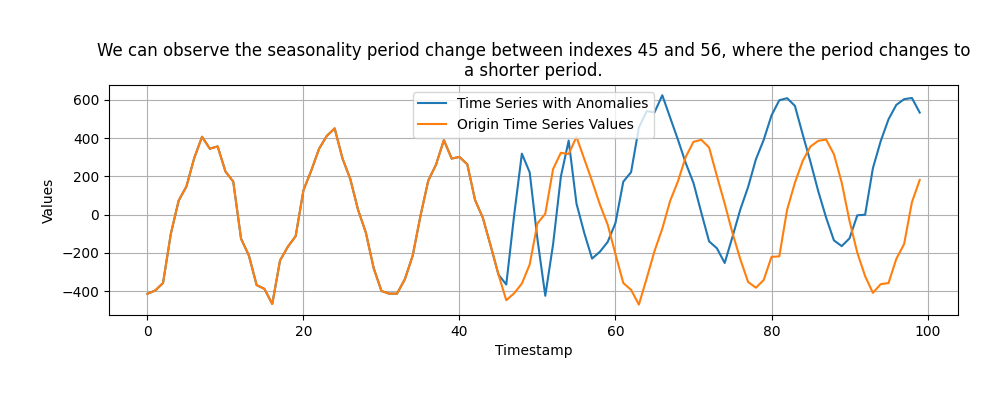}
\centering{(a) Period Change}
\end{minipage}
\begin{minipage}[t]{0.49\linewidth}
\includegraphics[width=\linewidth]{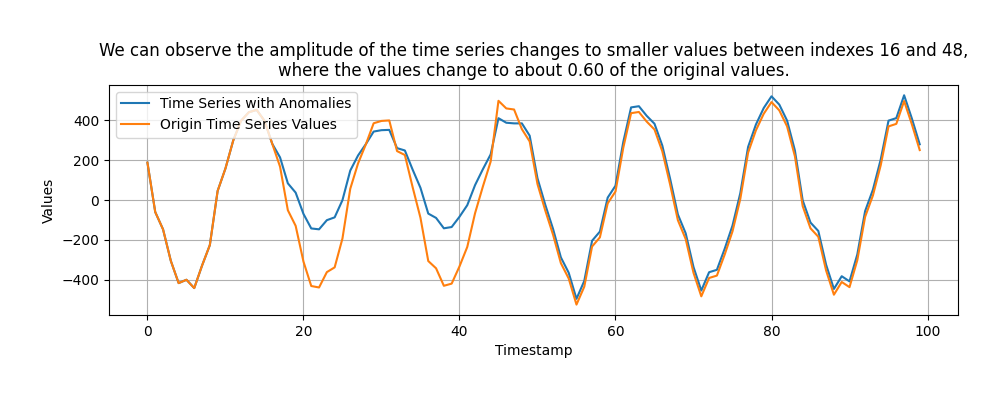}
\centering{(b) Amplitude Change}
\end{minipage}
    \caption{Examples for the time series with seasonality anomalies with automatically generated anomaly explanation.}
    \label{fig:example_anomaly_season}
\end{figure}

\begin{figure}
    \centering
\begin{minipage}[t]{0.49\linewidth}
\includegraphics[width=\linewidth]{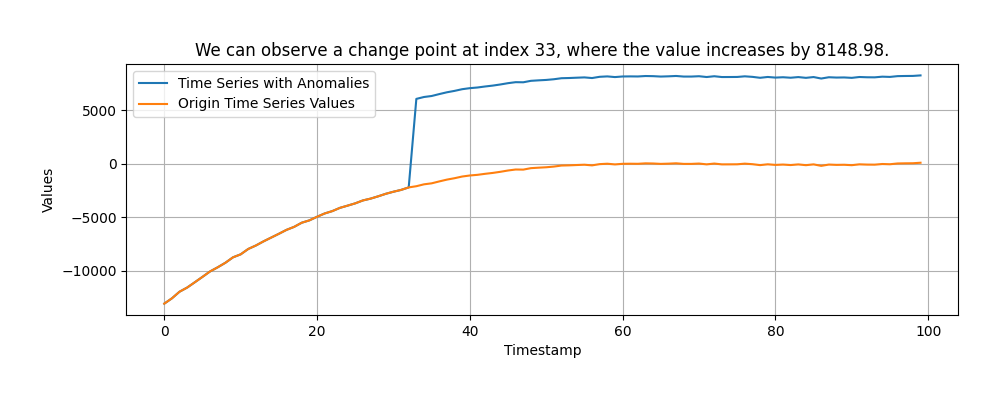}
\centering{(a) Trend Change}
\end{minipage}
\begin{minipage}[t]{0.49\linewidth}
\includegraphics[width=\linewidth]{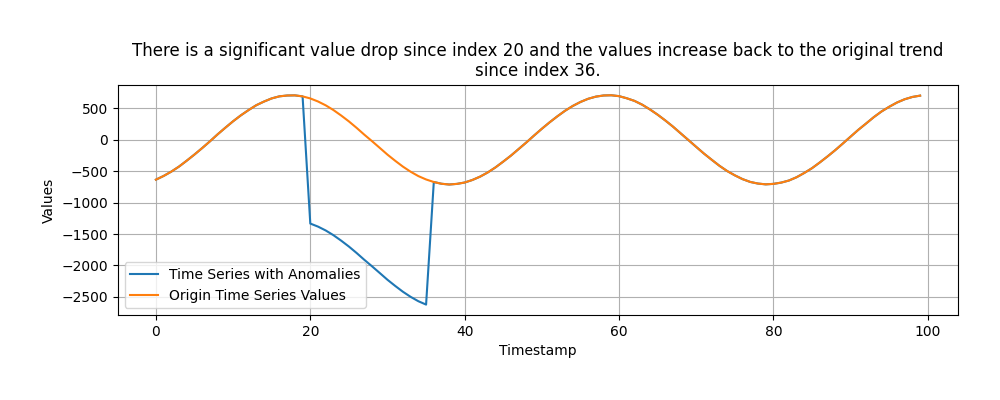}
\centering{(b) Trend break}
\end{minipage}
    \caption{Examples for the time series with trend anomalies with automatically generated anomaly explanation.}
    \label{fig:example_anomaly_trend}
\end{figure}

\begin{figure}
    \centering
\begin{minipage}[t]{0.49\linewidth}
\includegraphics[width=\linewidth]{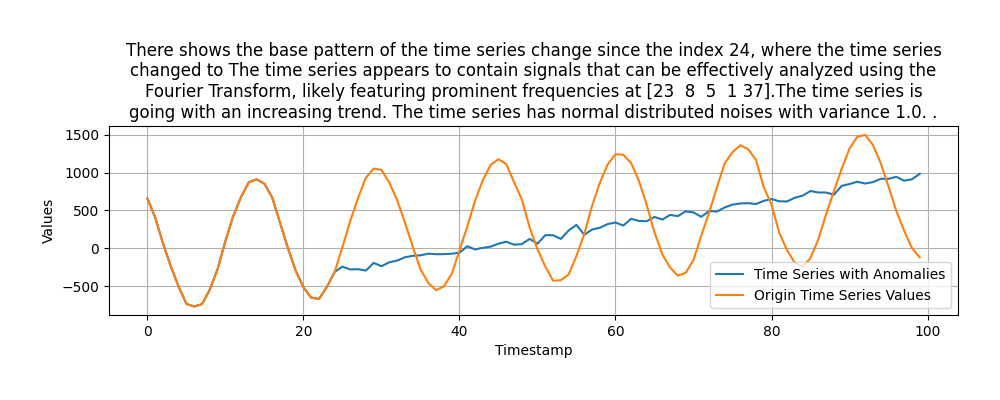}
\centering{(a) Shape Change}
\end{minipage}
\begin{minipage}[t]{0.49\linewidth}
\includegraphics[width=\linewidth]{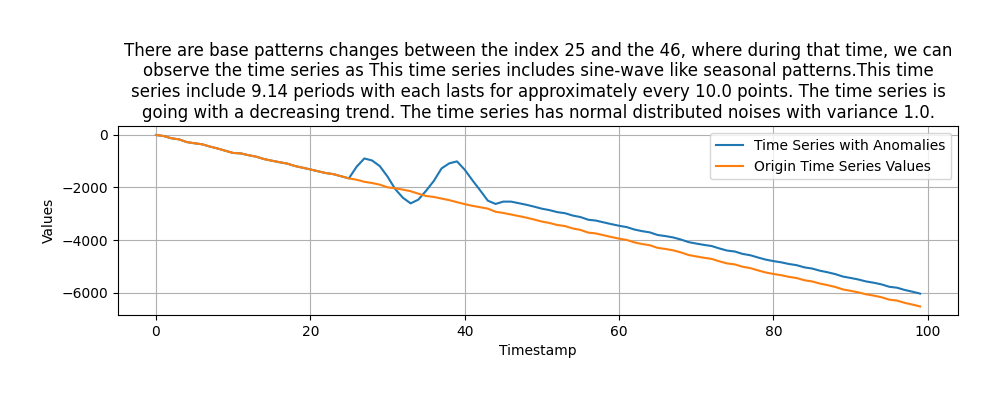}
\centering{(b) Shape Break}
\end{minipage}
    \caption{Examples for the time series with shape anomalies with automatically generated anomaly explanation.}
    \label{fig:example_anomaly_shape}
\end{figure}

\textbf{Examples for the Generated Samples}
Figure~\ref{fig:example_base_generator} shows examples of the base time series generation results and their corresponding automatically generated explanations. Figure~\ref{fig:example_anomaly_point} illustrates examples of global and local point anomalies. Figure~\ref{fig:example_anomaly_season} presents examples of seasonality anomalies. Figure~\ref{fig:example_anomaly_trend} displays examples of trend anomalies, and Figure~\ref{fig:example_anomaly_shape} demonstrates examples of shape anomalies. Note that all explanations are automatically generated by our template and will be further refined by GPT-4.

\textbf{Formalization to Instruction Dataset}
Specifically, a single data sample includes the time series values, labels for the anomalies, labels for the specific types of anomalies, an explanation for the base time series only, an explanation for the anomalies only, an explanation including both the base and anomaly explanations, and an explanation rewritten by the LLMs.
We construct datasets with 100, 500, 1000 and 2000 samples, with time series lengths of 180, 360, and 720. Although in real applications the total length of the time series can be quite long, we do not consider this due to the context window limitations of the LLMs. 
During instruction fine-tuning, the prompt we feed into LLaMA3 is the concatenation of the ideal explanation with the instruction prompt and the requirements prompt as described in Appendix~\ref{app:prompt_settings}. 

\textbf{Formalization to Evaluation Dataset}
We generated three datasets, each containing 100 samples, with time series lengths of 100, 200, and 400, respectively. Each dataset includes 20 samples for each of the following anomaly types: global point anomaly, local point anomaly, seasonality anomaly, trend anomaly, and shape anomaly. This distribution allows us to effectively evaluate model performance across different types of time series anomalies and various time series lengths. Additionally, the evaluation dataset lengths differ from those used to fine-tune LLaMA3 to prevent the model from learning any specific length-related patterns or tricks.

\textbf{Difference from \cite{lai2021revisiting}} Although we have followed the definition of anomaly types from this work, TTGenerator considers more detailed types for the base pattern, including additional trend and seasonality types, as well as more detailed anomaly types. Most importantly, this work did not provide automatic textual descriptions for the time series, which are essential for constructing an instruction dataset.

\begin{figure}[t]
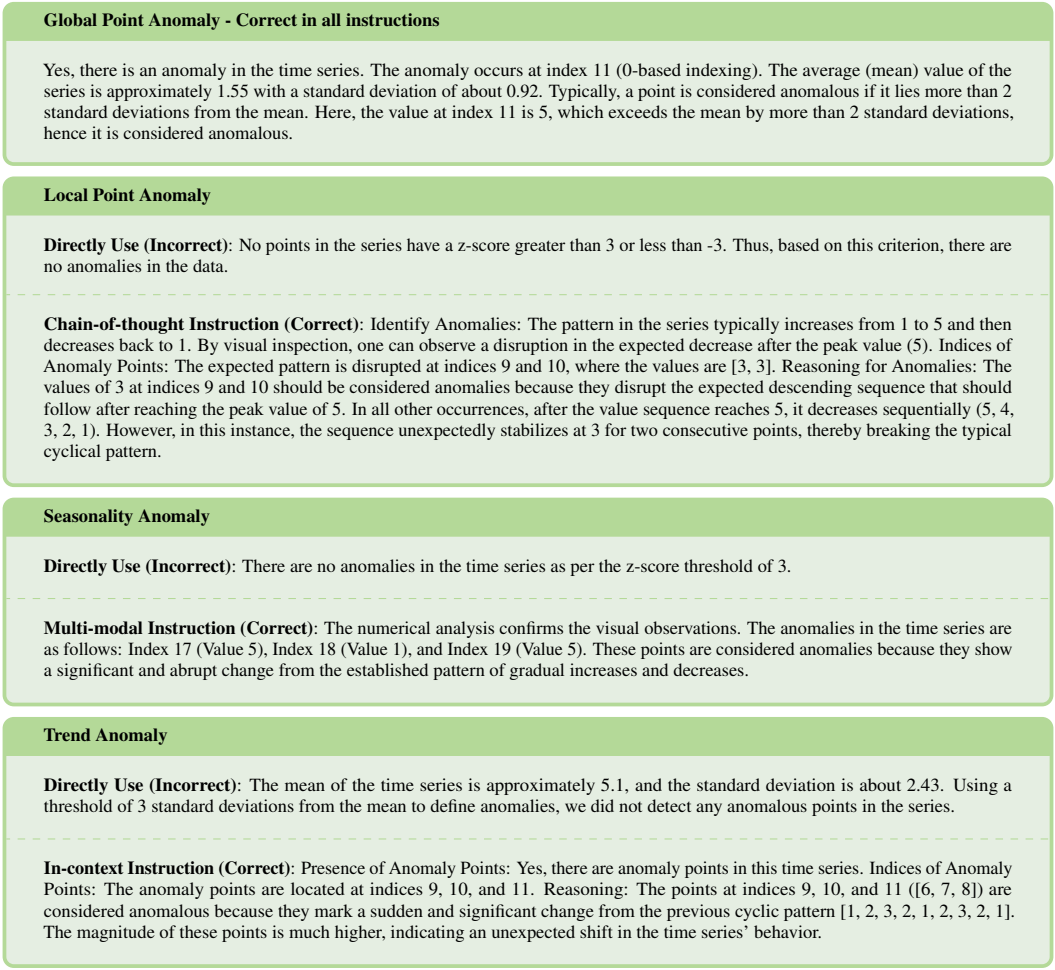

    \centering
    \scriptsize

\begin{tcolorbox}[colback=OliveGreen!10!white,colframe=YellowGreen!70!white,title=\textbf{Global Point Anomaly - Correct in all instructions},coltitle=black]
Yes, there is an anomaly in the time series. The anomaly occurs at index 11 (0-based indexing). The average (mean) value of the series is approximately 1.55 with a standard deviation of about 0.92. Typically, a point is considered anomalous if it lies more than 2 standard deviations from the mean. Here, the value at index 11 is 5, which exceeds the mean by more than 2 standard deviations, hence it is considered anomalous. 

\end{tcolorbox}

\begin{tcolorbox}[colback=OliveGreen!10!white,colframe=YellowGreen!70!white, title=\textbf{Local Point Anomaly}, coltitle=black]
\textbf{Directly Use (Incorrect)}: No points in the series have a z-score greater than 3 or less than -3. Thus, based on this criterion, there are no anomalies in the data.
\tcblower
\textbf{Chain-of-thought Instruction (Correct)}: Identify Anomalies:
The pattern in the series typically increases from 1 to 5 and then decreases back to 1. By visual inspection, one can observe a disruption in the expected decrease after the peak value (5).
Indices of Anomaly Points:
The expected pattern is disrupted at indices 9 and 10, where the values are [3, 3]. 
Reasoning for Anomalies:
The values of 3 at indices 9 and 10 should be considered anomalies because they disrupt the expected descending sequence that should follow after reaching the peak value of 5. In all other occurrences, after the value sequence reaches 5, it decreases sequentially (5, 4, 3, 2, 1). However, in this instance, the sequence unexpectedly stabilizes at 3 for two consecutive points, thereby breaking the typical cyclical pattern. 
\end{tcolorbox}

\begin{tcolorbox}[colback=OliveGreen!10!white,colframe=YellowGreen!70!white, title=\textbf{Seasonality Anomaly}, coltitle=black]
\textbf{Directly Use (Incorrect)}: There are no anomalies in the time series as per the z-score threshold of 3.
\tcblower
\textbf{Multi-modal Instruction (Correct)}: The numerical analysis confirms the visual observations. The anomalies in the time series are as follows: Index 17 (Value 5), Index 18 (Value 1), and Index 19 (Value 5). These points are considered anomalies because they show a significant and abrupt change from the established pattern of gradual increases and decreases. 

\end{tcolorbox}

\begin{tcolorbox}[colback=OliveGreen!10!white,colframe=YellowGreen!70!white, title=\textbf{Trend Anomaly}, coltitle=black]
\textbf{Directly Use (Incorrect)}: The mean of the time series is approximately 5.1, and the standard deviation is about 2.43. Using a threshold of 3 standard deviations from the mean to define anomalies, we did not detect any anomalous points in the series.
\tcblower
\textbf{In-context Instruction (Correct)}: 
Presence of Anomaly Points: Yes, there are anomaly points in this time series. 
Indices of Anomaly Points: The anomaly points are located at indices 9, 10, and 11. 
Reasoning: The points at indices 9, 10, and 11 ([6, 7, 8]) are considered anomalous because they mark a sudden and significant change from the previous cyclic pattern [1, 2, 3, 2, 1, 2, 3, 2, 1]. The magnitude of these points is much higher, indicating an unexpected shift in the time series' behavior.
\end{tcolorbox}
\caption{More examples responses of GPT4 on the trial cases for other anomaly types. 
}
\label{fig:GPT4_responses}
\end{figure}

\begin{figure}[t]
    \centering
    \scriptsize

\begin{tcolorbox}[colback=Dandelion!10!white,colframe=Dandelion!70!white,title=\textbf{Global Point Anomaly},coltitle=black]
\textbf{Directly Use (Correct)} Yes, there is at least one anomaly point in the time series. The index of the anomaly point is 11. Reasoning behind the anomaly: The value 5 at index 11 is considered an anomaly because it is significantly different from the rest of the values in the series. The majority of the values in the series are either 1 or 2, with a few occurrences of 1 and 2. The value 5 is significantly higher than the other values and does not follow the pattern of the series.

\textbf{Multi-modal Instruction (Correct)} Presence of anomaly points: Yes. Indices of anomaly points: 11 (index 11 corresponds to the value 5).
Reasoning: The value 5 at index 11 is significantly higher than the rest of the values in the series, which are mostly in the range of 1 and 2. This outlier value stands out as an anomaly due to its unusual magnitude and deviation from the overall pattern of the series.

\textbf{Multi-modal Instruction (Incorrect)} Yes, there are several anomaly points in this time series. The indices of the anomaly points are:
Index 11 (value: 5): This value is significantly higher than the surrounding values, which are mostly in the range of 1-2. The sudden jump to 5 is unusual and may indicate an anomaly.
Index 20 (value: 2): This value is higher than the surrounding values, which are mostly 1. The sudden increase in value may indicate an anomaly.

\end{tcolorbox}

\begin{tcolorbox}[colback=Dandelion!10!white,colframe=Dandelion!70!white, title=\textbf{Local Point Anomaly}, coltitle=black]
\textbf{Directly Use (Incorrect)}: Presence of anomaly points: Yes. Indices of anomaly points: 4, 12. Reasoning behind these points being considered anomalies:
- Index 4: The value 5 has a Z-score of 1.29, which is greater than 2 standard deviations away from the mean. This indicates that this data point is an anomaly.
- Index 12: The value 5 has a Z-score of 1.29, which is greater than 2 standard deviations away from the mean. This indicates that this data point is an anomaly.

\textbf{In-context Instruction (Incorrect)} Yes, there are several anomaly points in this time series. The indices of the anomaly points are:
Index 9 (value: 3): This value is different from the surrounding values, which are mostly increasing. The sudden decrease in value may indicate an anomaly.
Index 10 (value: 3): This value is the same as the previous value, which is unusual in a time series that is generally increasing.
Index 17 (value: 2): This value is significantly lower than the surrounding values, which are mostly increasing. The sudden decrease in value may indicate an anomaly.

\end{tcolorbox}

\begin{tcolorbox}[colback=Dandelion!10!white,colframe=Dandelion!70!white, title=\textbf{Seasonality Anomaly}, coltitle=black]
\textbf{Directly Use (Incorrect)}:  Yes, there are anomalies in the time series. Indices of anomaly points:
Index 0: The value 1 is an anomaly because it's significantly lower than the previous value (0) and the overall trend of the series.
Index 14: The value 5 is an anomaly because it's significantly higher than the previous value (1) and the overall trend of the series.
Index 17: The value 5 is an anomaly because it's significantly higher than the previous value (1) and the overall trend of the series.

\textbf{Chain-of-thought Instruction (Incorrect)} Yes, there are anomaly points in this time series. The anomaly points are at indices [12] and [17]. Reasoning behind these points being considered anomalies: The values at indices [12] and [17] are 5, which is significantly higher than the previous and subsequent values. The time series has a cyclical pattern with decreasing values, and the sudden increase at indices [12] and [17] breaks this pattern, making them anomalies.
\end{tcolorbox}

\begin{tcolorbox}[colback=Dandelion!10!white,colframe=Dandelion!70!white, title=\textbf{Trend Anomaly}, coltitle=black]
\textbf{Directly Use (Partially Correct)}: Yes, there are anomalies in the time series. The indices of the anomaly points are 9, 10, 11, and 12. Reasoning behind the anomaly points being considered anomalies: The points at indices 9, 10, and 11 (values 6, 7, and 8) are significantly higher than the rest of the time series, which consists mainly of values ranging from 1 to 3. These higher values are likely anomalies because they deviate significantly from the overall pattern of the time series. 

\textbf{Multi-modal Instruction (Correct)}: Presence of anomaly points: Yes
Indices of anomaly points: 9, 10, 11
Reasoning behind these points being considered anomalies: The time series can be visualized as a wave-like pattern with oscillations between 1, 2, and 3. The anomaly points at indices 9, 10, and 11 appear as a sudden spike in the wave pattern, making them stand out as anomalies.

\end{tcolorbox}
\caption{More example responses of LLaMA3 on the trial cases for other anomaly types. 
}
\label{fig:LLaMA3_responses}
\end{figure}

\begin{figure*}[t]
\centering
\small
\begin{minipage}[t]{0.3\linewidth}
\includegraphics[width=\linewidth]{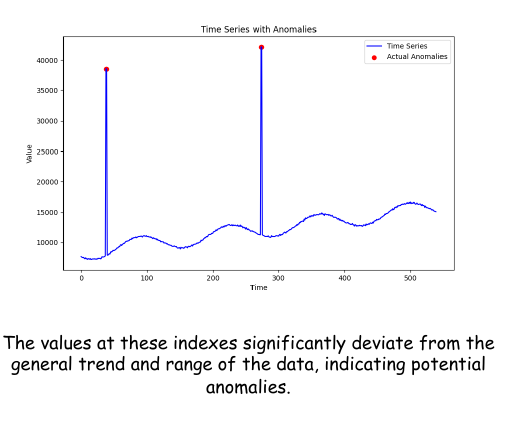}
\centering{(a) YAHOO - Good}
\end{minipage}
\begin{minipage}[t]{0.3\linewidth}
\includegraphics[width=\linewidth]{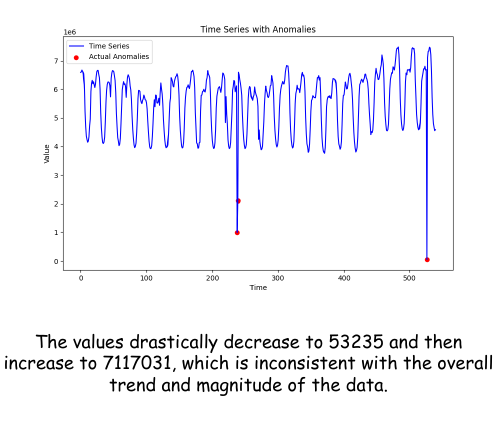}
\centering{(b) YAHOO - Bad}
\end{minipage}
\begin{minipage}[t]{0.3\linewidth}
\includegraphics[width=\linewidth]{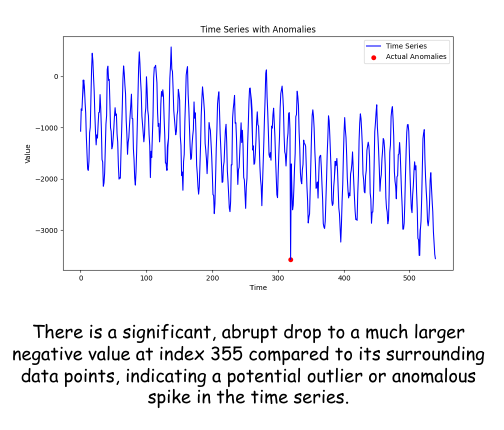}
\centering{(c) YAHOO - Hallucination}
\end{minipage}
\begin{minipage}[t]{0.3\linewidth}
\includegraphics[width=\linewidth]{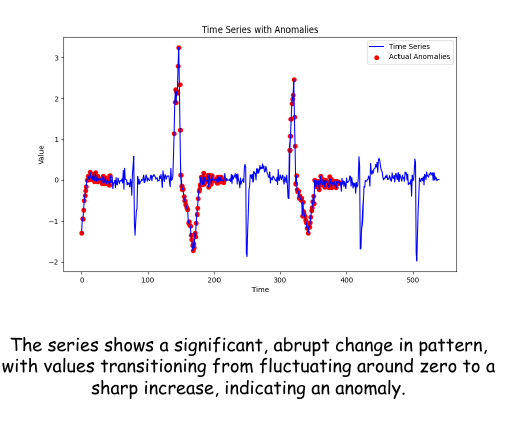}
\centering{(a) ECG - Good}
\end{minipage}
\begin{minipage}[t]{0.3\linewidth}
\includegraphics[width=\linewidth]{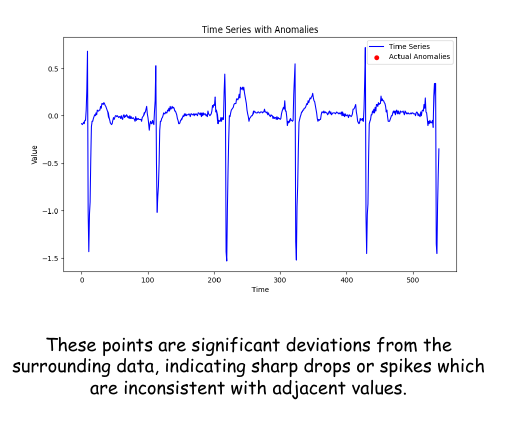}
\centering{(b) ECG - Bad}
\end{minipage}
\begin{minipage}[t]{0.3\linewidth}
\includegraphics[width=\linewidth]{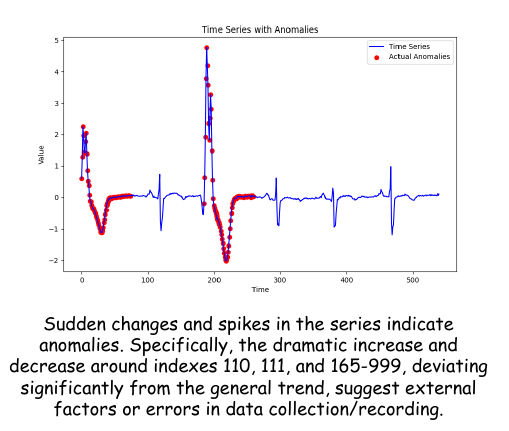}
\centering{(c) ECG - Hallucination}
\end{minipage}
\begin{minipage}[t]{0.3\linewidth}
\includegraphics[width=\linewidth]{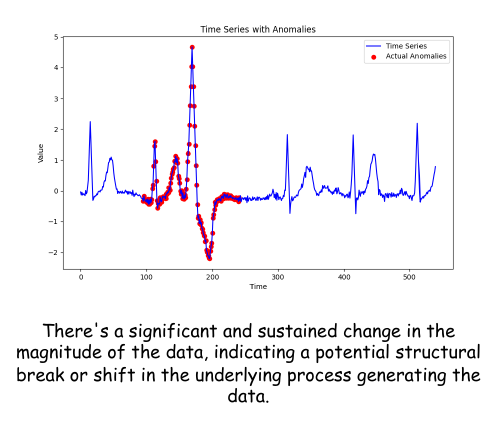}
\centering{(a) SVDB - Good}
\end{minipage}
\begin{minipage}[t]{0.3\linewidth}
\includegraphics[width=\linewidth]{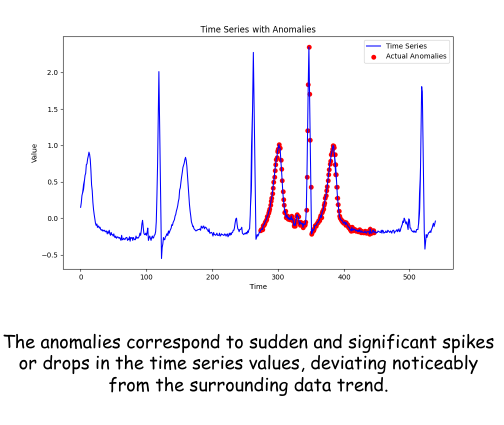}
\centering{(b) SVDB - Bad}
\end{minipage}
\begin{minipage}[t]{0.3\linewidth}
\includegraphics[width=\linewidth]{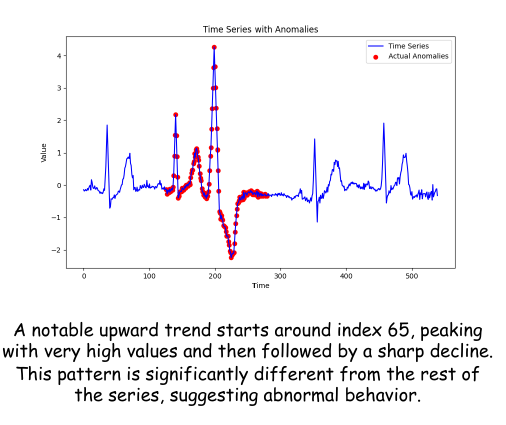}
\centering{(c) SVDB - Hallucination}
\end{minipage}
\caption{Examples for a) good, b) bad, and c) hallucinated explanation by GPT-4 on YAHOO, ECG, and SVDB datasets.}
\label{fig:example_explanation_all}
\end{figure*}

\begin{figure}[t]
\centering
\small
\begin{minipage}[t]{0.23\linewidth}
\includegraphics[width=\linewidth]{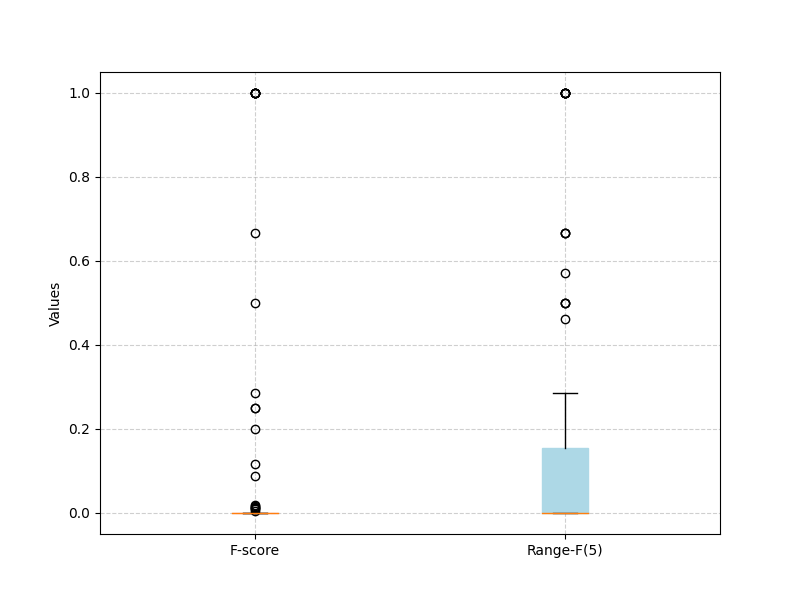}
\centering{(a) YAHOO}
\end{minipage}
\begin{minipage}[t]{0.23\linewidth}
\includegraphics[width=\linewidth]{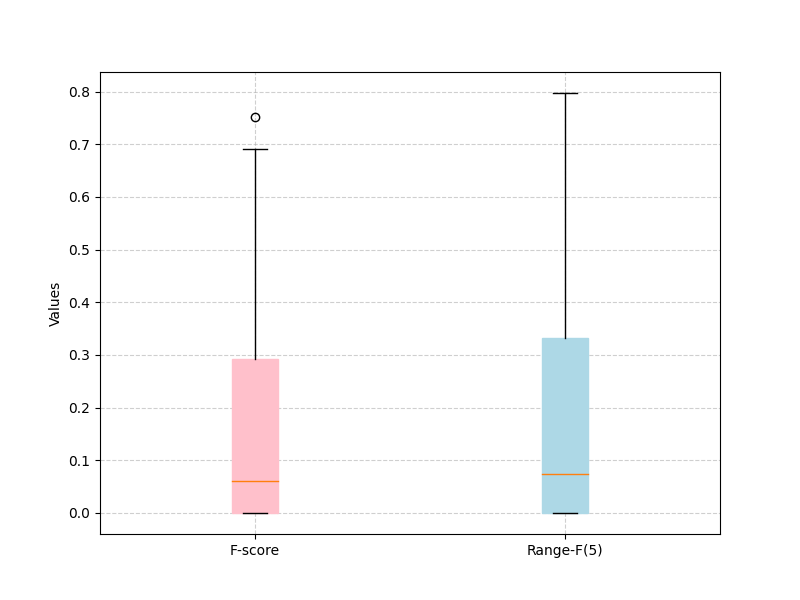}
\centering{(b) ECG}
\end{minipage}
\begin{minipage}[t]{0.23\linewidth}
\includegraphics[width=\linewidth]{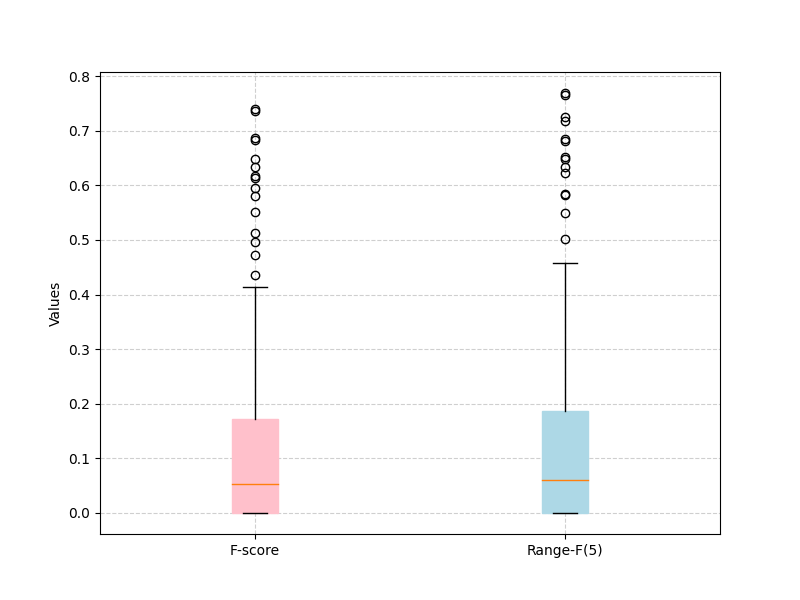}
\centering{(c) SVDB}
\end{minipage}
\begin{minipage}[t]{0.23\linewidth}
\includegraphics[width=\linewidth]{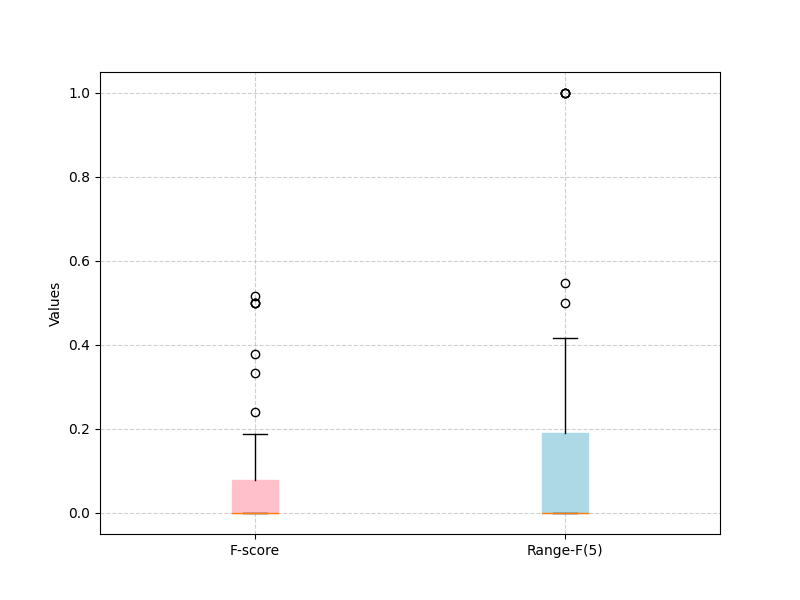}
\centering{(d) IOPS}
\end{minipage}
\caption{The distribution over F-score and Range-F on the four datasets by GPT-4.}
\label{fig:gpt4_f-score_distribution}
\end{figure}

\begin{figure}[t]
\centering
\small
\begin{minipage}[t]{0.23\linewidth}
\includegraphics[width=\linewidth]{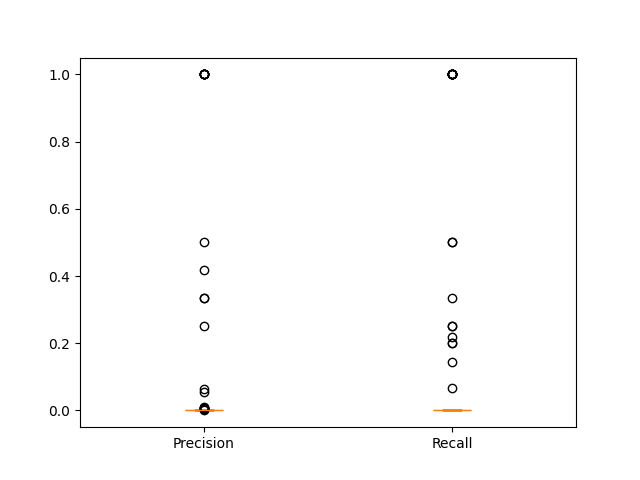}
\centering{(a) YAHOO}
\end{minipage}
\begin{minipage}[t]{0.23\linewidth}
\includegraphics[width=\linewidth]{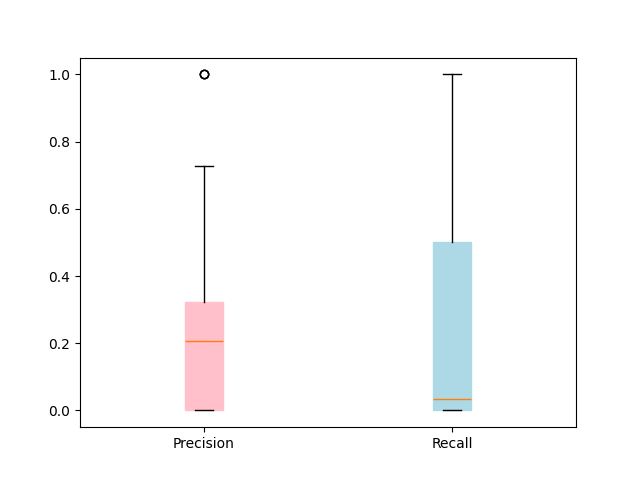}
\centering{(b) ECG}
\end{minipage}
\begin{minipage}[t]{0.23\linewidth}
\includegraphics[width=\linewidth]{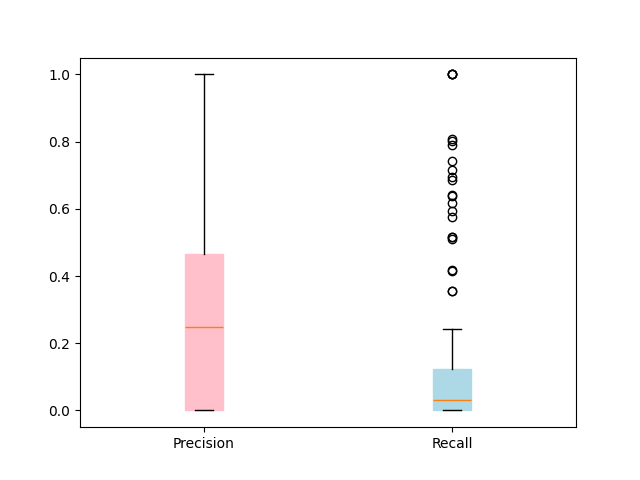}
\centering{(c) SVDB}
\end{minipage}
\begin{minipage}[t]{0.23\linewidth}
\includegraphics[width=\linewidth]{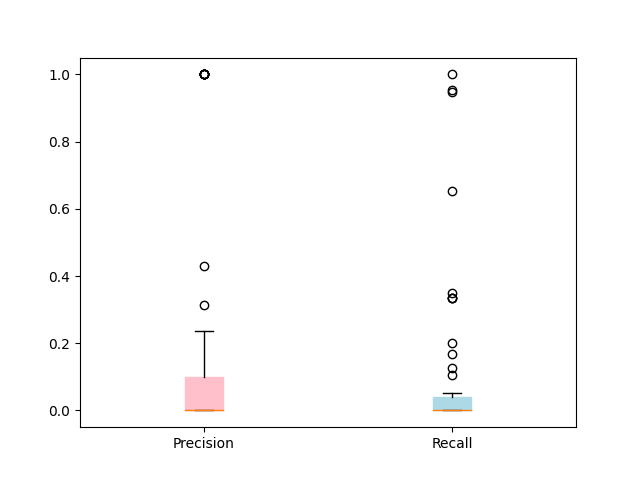}
\centering{(d) IOPS}
\end{minipage}
\caption{The distribution over F-score and Range-F on the four datasets by GPT-4.}
\label{fig:gpt4_precision_distribution}
\end{figure}

\begin{figure}[t]
\centering
\small
\begin{minipage}[t]{0.23\linewidth}
\includegraphics[width=\linewidth]{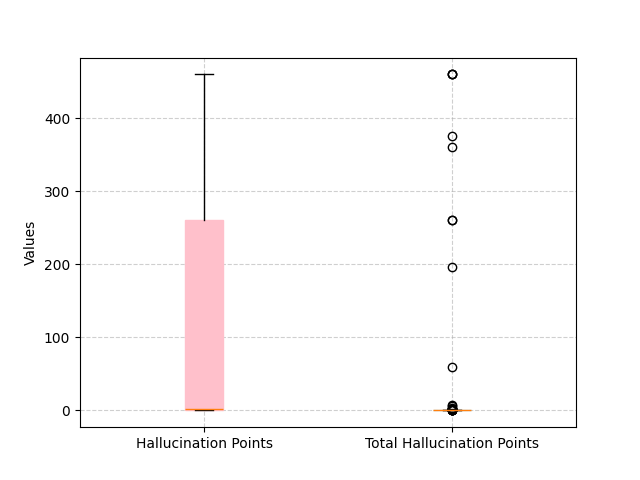}
\centering{(a) YAHOO}
\end{minipage}
\begin{minipage}[t]{0.23\linewidth}
\includegraphics[width=\linewidth]{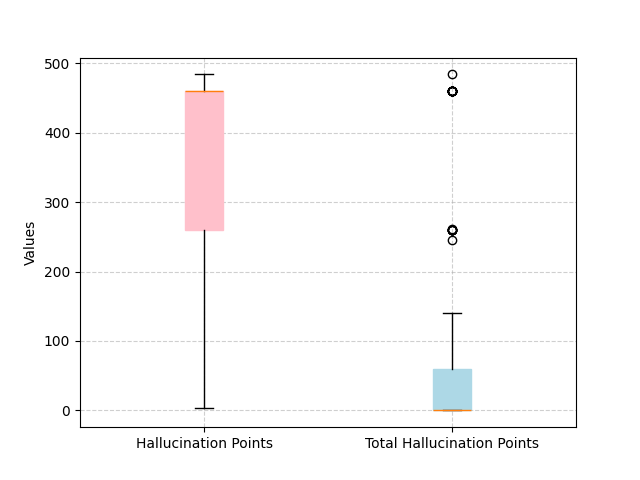}
\centering{(b) ECG}
\end{minipage}
\begin{minipage}[t]{0.23\linewidth}
\includegraphics[width=\linewidth]{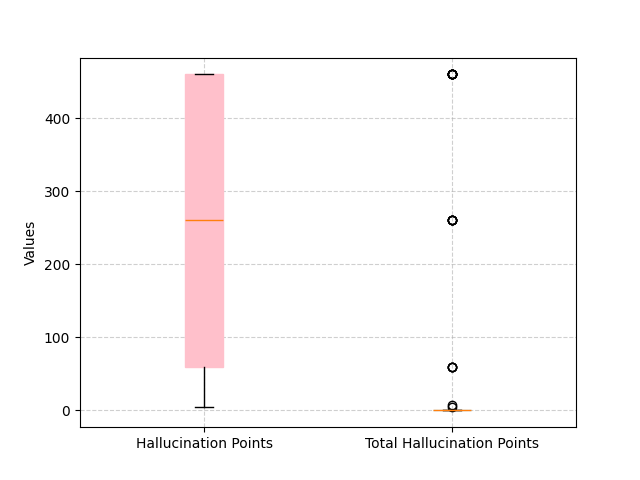}
\centering{(c) SVDB}
\end{minipage}
\begin{minipage}[t]{0.23\linewidth}
\includegraphics[width=\linewidth]{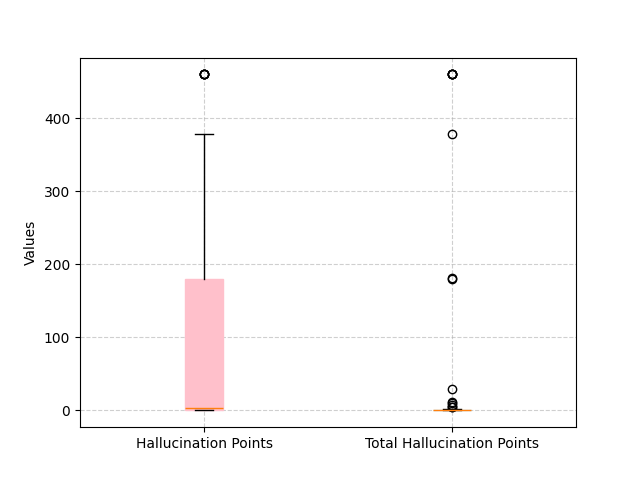}
\centering{(d) IOPS}
\end{minipage}
\caption{The distribution of left: the hallucination points over the segments that the model has hallucination and right: the total hallucination points over all segments.}
\label{fig:hallucination_distribution}
\end{figure}

\subsection{LLM Settings}
\label{app:llm_settings}
For both models, we employ a rerun strategy: if the model fails to provide the required JSON-formatted response, we automatically rerun the code until the response adheres to the specified format. If the model fails more than five trials, we return the default response as \{\texttt{anomaly}: [], \texttt{reason}: ""\}.

\paragraph{GPT-4 Settings}
The GPT-4 version we used is `GPT-4-0125-preview`\footnote{\url{https://openai.com/pricing}}. Specifically, we employ LangChain\footnote{\url{https://www.langchain.com/}} to facilitate the integration of prompts with the OpenAI API and to parse the output into JSON format for easier evaluation. We use the default parameters in generating the responses. 

\paragraph{LLaMA3 Settings}
To obtain the inference results for the original LLaMA3, we utilized Groq's API services\footnote{\url{https://groq.com/}}. All results were generated using LLaMA3-8B due to computational resource limitations.
Our fine-tuning is based on Meta-Llama-3-8B-Instruct\footnote{\url{https://huggingface.co/meta-llama/Meta-Llama-3-8B-Instruct}}. We used parameter-efficient fine-tuning approaches, specifically LoRA~\citep{hu2021lora}, with Hugging Face's PEFT packages\footnote{\url{https://huggingface.co/docs/peft/index}}. The settings for LoRA are as follows: we fine-tuned all linear layers in the transformers, with the LoRA rank set to 16, LoRA alpha set to 64, and a dropout rate of 0.1.
The training arguments are: gradient accumulation for 4 steps, using the \texttt{paged\_adamw\_8bit} optimizer, a learning rate of 2e-4 with a cosine learning rate scheduler, a warmup ratio of 0.05, a max gradient norm of 0.3, \texttt{fp16} set to \texttt{True}, and \texttt{group\_by\_length} set to \texttt{True}. We trained for 1 epoch, as we observed a drop in model performance with additional epochs.
During inference, we set the \texttt{max\_new\_tokens} parameter to 512 due to computational limitations (processing 100 samples with this parameter set to 2048 takes approximately 5 hours). We used the default generation strategy, which is greedy search. 
The instruction dataset is configured to contain 1000 samples generated by TTGenerator, with a mix of time series lengths of 180, 360, and 720, and includes a variety of five types of time series anomalies or no anomaly. By default, we use the general instruction prompt to formalize the text for fine-tuning LLaMA3.

\subsection{Computation Resources}
\label{app:compute_resource}
The experiments are conducted on two NVIDIA H100 PCIe 80G GPUs. Fine-tuning LLaMA-3-8B on 2000 datasets with the time series length as 360 takes approximately 2 hours and requires about 130 GB of memory. Inference requires about 15 GB of memory on a single GPU and takes approximately 1 hour for 100 data samples.






\section{Complementary Results}

\subsection{Example Responses Given Different Prompt Strategies On Trial Cases}
\label{app:exp_trial_examples}
Figure~\ref{fig:GPT4_responses} and Figure~\ref{fig:LLaMA3_responses} present additional examples of responses generated by GPT-4 and LLaMA3 on the trial cases. The responses produced by LLaMA3 are significantly more unstable compared to those from GPT-4. For instance, when identifying global point anomalies, LLaMA3 can yield different outputs for the same prompt. Although LLaMA3 does not benefit much from prompt engineering, differences can still be observed before and after applying prompting strategies. In some cases, LLaMA3 provides responses that closely match the ideal answers.

\subsection{More results on GPT-4}
\label{app:evaluation_gpt4}

\begin{figure}[ht]
    \centering
    \includegraphics[width=0.3\linewidth]{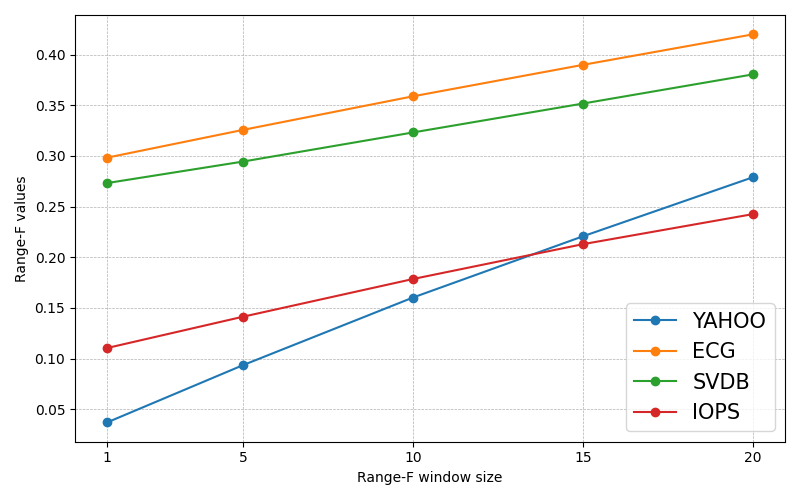}
    \caption{Results with different Range-F window size.}
    \label{fig:range_f_window_size}
\end{figure}

\textbf{Performance under different Range-F window size}\quad
The evaluation metric range-F score evaluates the model's precision in detecting anomalies. Specifically, it questions how closely the model's predicted anomaly positions align with the actual anomaly positions. Figure~\ref{fig:range_f_window_size} illustrates the variation of the F-score as the range-F window size changes across different benchmark datasets. Generally, enlarging the window size enhances the model's performance, with notable improvements observed in the YAHOO dataset. This suggests that, compared to other datasets, the model's predictions for anomaly positions in the YAHOO dataset are significantly closer to their actual locations.

\textbf{More results in evaluation metrics}\quad
The distribution over F-score and Range-F on the four benchmark datasets is shown in Figure~\ref{fig:gpt4_f-score_distribution}. 
We further explore the precision and recall across various datasets. The outcomes are detailed in Table~\ref{tab:precision_recall}. Generally, we observe that recall exceeds precision, indicating a model propensity to identify a broader array of potential anomalies, albeit with lower precision. Conversely, for the SVDB dataset, precision surpasses recall, suggesting a more selective approach by the model in flagging anomalies. This phenomenon may be attributed to the anomaly composition within the datasets, as elaborated in Table~\ref{tab:dataset_details} found in Appendix \ref{app:benchmark_dataset_settings}. Specifically, SVDB exhibits a higher ratio of anomalous points. 
\begin{table}[t]
    \centering
    \tiny
        \caption{Precision and Recall on the four datasets by GPT4}
    \begin{tabular}{ccccc}
    \toprule
        Metrics & YAHOO & ECG & SVDB & IOPS \\ \midrule
        Precision & 0.0127 & 0.2654 & 0.3024 & 0.0880 \\ 
        Recall & 0.0520 & 0.3225 & 0.2408 & 0.1214 \\
        \bottomrule
    \end{tabular}
    \label{tab:precision_recall}
\end{table}
The distribution of the precision and recall over the four datasets is are shown in Figure~\ref{fig:gpt4_precision_distribution}.

\textbf{More Examples for the Explanation on Different Datasets}
Figure~\ref{fig:example_explanation_all} shows additional examples of explanations provided by GPT-4 on different datasets. We observe that for the ECG and SVDB datasets, GPT-4 struggles to accurately interpret the ECG signals, occasionally mistaking normal spike and dip patterns for anomalies. However, when the changes are obvious, GPT-4 is able to detect such pattern changes. The issues with the YAHOO and IOPS datasets arise when GPT-4 hallucinates in terms of direction, indexes, or change values.

\textbf{More analysis for the hallucination in indices}
Figure~\ref{fig:hallucination_distribution} illustrates the distribution of hallucinations in indices across the four datasets. In each figure, the left part shows the distribution of the number of hallucination indices within the time series segments where GPT-4 hallucinated in the identified anomalous indices. The right part displays the distribution of the number of hallucination points across all 100 segments for each dataset. The results indicate that GPT-4 does not hallucinate in indices in most time series segments; however, when the model does hallucinate, the number of hallucination points tends to be large.
\begin{table}[t]
    \centering
    \tiny
    \caption{Performance after filtering the hallucination segments.}
    \begin{tabular}{ccc}
    \toprule
        Dataset & F-score & Range-F  \\ \midrule
        YAHOO & 0.0440(\textcolor{green}{$\uparrow$}0.0236) & 0.1027(\textcolor{green}{$\uparrow$}0.0091) \\
        ECG & 0.1537(\textcolor{red}{$\downarrow$}0.1374) & 0.1736(\textcolor{red}{$\downarrow$}0.1522) \\
        SVDB & 0.1683(\textcolor{red}{$\downarrow$}0.0998) & 0.1850(\textcolor{red}{$\downarrow$}0.1002) \\
        IOPS & 0.0161(\textcolor{red}{$\downarrow$}0.0859) & 0.0245(\textcolor{red}{$\downarrow$}0.1169) \\ \bottomrule
    \end{tabular}
    \label{tab:hallucination_remove_stats}
\end{table}

Subsequently, we evaluate the model's performance after excluding segments with hallucinations, with the results detailed in Table~\ref{tab:hallucination_remove_stats}. Ideally, one would expect an improvement in performance after removing these affected segments. Surprisingly, except for the YAHOO dataset, the remaining three datasets exhibited a notable decrease in performance. This suggests a tendency for the model to hallucinate more frequently in segments with a higher proportion of anomalies. Moreover, it indicates that, for these segments, the model attempts to identify additional positions, even though some may not align with the actual segments. 
\begin{table}[h]
    \centering
    \tiny
    \caption{Evaluation Metrics for GPT-4 on Synthesized Dataset}
    \begin{tabular}{cccccc}
    \toprule
    Timeseries Length & Anomaly Type & F-score & Range-F & Precision & Recall \\
    \midrule
    \multirow{5}{*}{100} 
    & Context Seasonal & 0.2351 & 0.3180 & 0.3992 & 0.1667 \\
    & Context Shape & 0.1993 & 0.3140 & 0.1601 & 0.2638 \\
    & Context Trend & 0.2688 & 0.4232 & 0.1675 & 0.6800 \\
    & Point Global & 0.4576 & 0.7000 & 0.4426 & 0.4737 \\
    & Point Local & 0.3934 & 0.6230 & 0.4068 & 0.3810 \\
    \midrule
    \multirow{5}{*}{200} 
    & Context Seasonal & 0.1524 & 0.1992 & 0.2480 & 0.1100 \\
    & Context Shape & 0.0869 & 0.1624 & 0.0616 & 0.1473 \\
    & Context Trend & 0.1263 & 0.2460 & 0.0944 & 0.1909 \\
    & Point Global & 0.2362 & 0.4531 & 0.2381 & 0.2344 \\
    & Point Local & 0.1477 & 0.3399 & 0.1294 & 0.1719 \\
    \midrule
    \multirow{5}{*}{400} 
    & Context Seasonal & 0.2444 & 0.2757 & 0.4332 & 0.1702 \\
    & Context Shape & 0.3202 & 0.3509 & 0.2226 & 0.5703 \\
    & Context Trend & 0.1305 & 0.1676 & 0.0826 & 0.3101 \\
    & Point Global & 0.2000 & 0.2600 & 0.2564 & 0.1639 \\
    & Point Local & 0.0656 & 0.1475 & 0.0784 & 0.0563 \\
    \bottomrule
    \end{tabular}
    \label{tab:gpt4_multimodal_evaluation}
\end{table}

\textbf{Performance on Synthesized Dataset} We further evaluate GPT-4 on our synthesized dataset with different time series anomaly types and time series lengths, the results for F-score, Range-F, precision, and recall can be found in Table~\ref{tab:gpt4_multimodal_evaluation}. 

We observe that, generally, GPT-4 performs better on point-aware anomalies than on context-aware anomalies. Additionally, the overall performance decreases as the time series length increases.
Comparing global point anomalies and local point anomalies, GPT-4 performs better on global point anomalies, which is consistent to the results on trial cases. Within context-aware anomalies, when the time series length is short, GPT-4 has more difficulty identifying shape anomalies compared to the other two types of context-aware anomalies. However, as the time series length increases, the shape anomalies become more apparent, allowing GPT-4 to perform better on these anomalies.

\begin{table}[h]
    \centering
    \scriptsize
    \caption{Hallucination statistics for each time series length by GPT-4}
    \begin{tabular}{cccc}
    \toprule
    Time Series Length & Sum of Count & Average of Mean & Average of Median \\
    \midrule
    100 & 11 & 1.466666 & 1.2 \\
    200 & 18 & 41.896666 & 44.5 \\
    400 & 35 & 145.202784 & 143.0 \\
    \bottomrule
    \end{tabular}
    \label{tab:gpt4_datasize_summary}
\end{table}

\begin{table}[h]
    \centering
    \scriptsize
    \caption{Hallucination statistics for each anomaly type by GPT-4}
    \begin{tabular}{cccc}
    \toprule
    Anomaly Type & Sum of Count & Average of Mean & Average of Median \\
    \midrule
    Seasonal & 12 & 85.833333 & 71.666667 \\
    Shape & 17 & 142.870370 & 170.833333 \\
    Trend & 15 & 83.339016 & 69.666667 \\
    Global Point & 10 & 1.190476 & 1.333333 \\
    Local Point & 10 & 1.041667 & 1.0 \\
    \bottomrule
    \end{tabular}
    \label{tab:gpt4_anomaly_type_summary}
\end{table}

Tables~\ref{tab:gpt4_datasize_summary} and \ref{tab:gpt4_anomaly_type_summary} present the statistics of hallucinations in indices on the synthetic dataset for different time series lengths and different types of time series anomalies.

From the perspective of time series length, we observe that hallucinations in indices become more significant as the time series length increases. The average number of hallucination points is small for shorter time series, with fewer than 2 points for datasets with a length of 100. However, for datasets with a length of 400, the problem becomes more severe, with the average number of hallucination points increasing to around 150.

From the perspective of anomaly type, GPT-4 tends to hallucinate more on context-aware anomalies, while it rarely hallucinates indices for global point and local point anomalies. Among the context-aware anomalies, GPT-4 is more prone to hallucinating indices in shape anomalies compared to the other two types of context-aware anomalies.

\subsection{More results on LLaMA3}
\label{app:llama3_results}

\textbf{Experiments on Benchmark Datasets} Due to the very limited context window of 8K tokens in LLaMA3, we attempted evaluation on LLaMA3 using four benchmark datasets. However, for the ECG and SVDB datasets, which contain about 30\% anomalies, LLaMA3 often failed to provide complete responses in many trials. As a result, we have not included the results for these four datasets in the main context. On the other hand, for the YAHOO dataset, the model was able to provide more complete responses. Therefore, we report the results for the YAHOO dataset in table~\ref{tab:llama3_yahoo_results}. We observe that after fine-tuning, LLaMA3 slightly outperforms GPT-4 on the YAHOO dataset.
\begin{table}[h]
    \centering
    \scriptsize
\caption{LLaMA3's performance on YAHOO dataset.}
    \begin{tabular}{ccccc}
    \toprule
        Metrics & F-score & Range-F & Precision & Recall \\ \midrule
        Original & 0.0181 & 0.1080 & 0.0096 & 0.1651 \\ 
        Fine-tuned & 0.0397 & 0.1121 & 0.0419 & 0.0377 \\
        \bottomrule
    \end{tabular}

    \label{tab:llama3_yahoo_results}
\end{table}

\begin{table}[t]
    \centering
    \scriptsize
    \caption{Statistics of the hallucinated indices of LLaMA3 by Anomaly Type}
    \begin{tabular}{ccccccc}
    \toprule
    \multirow{2}{*}{Anomaly Type} & \multicolumn{3}{c}{Original LLaMA3} & \multicolumn{3}{c}{Fine-tuned LLaMA3} \\
     & \# Segments & Mean & Median & \# Segments & Mean & Median \\
    \midrule
    Global Point & 38 & 204.1 & 103.3 & 23 & 156.6 & 152.5 \\
    Local Point & 38 & 211.2 & 58.7 & 24 & 140.9 & 99.0 \\
    Seasonal & 43 & 216.3 & 106.7 & 25 & 170.2 & 147.7 \\
    Shape & 44 & 200.9 & 123.8 & 29 & 240.3 & 257.7 \\
    Trend & 34 & 259.9 & 155.7 & 21 & 222.7 & 194.3 \\
    \bottomrule
    \end{tabular}
    \label{tab:summary_statistics_anomaly_type_llama3}
\end{table}

\textbf{More results on the Hallucination in Indices}
Table~\ref{tab:summary_statistics_anomaly_type_llama3} presents the details of hallucinated indices across different anomaly types for both the original LLaMA3 and the fine-tuned LLaMA3 models. Generally, the number of hallucinated indices is lower for point-aware anomalies. However, compared to GPT-4, LLaMA3 exhibits a significantly higher number of hallucinated indices for point-aware anomalies. Additionally, we observe a decrease in the number of hallucinated segments after fine-tuning.

\textbf{For the performance under different instruction dataset settings} 
We evaluate the impact of sample size for fine-tuning on an instruction dataset with 100, 500, 1000, and 2000 samples. Interestingly, fine-tuning with 2000 samples generally results in worse performance compared to using fewer samples. In some cases, fine-tuning with 500 samples outperforms models fine-tuned with more data. Overall, fine-tuning with 1000 samples yields the best performance in most cases, so we use this sample size by default for our fine-tuning process.
Regarding the time series length for fine-tuning, we find that fine-tuning on varying lengths (i.e., for each sample, randomly selecting the time series length in the range of 180, 360, and 720) consistently provides the best performance across different settings. Another observation is that when we fine-tune our models on datasets with time series lengths of 180, 360, and 720, and evaluate them on time series lengths of 100, 200, and 400, the reduction in hallucinated indices as the time series length decreases may be due to the fact that the evaluation lengths are closer to the lengths used during fine-tuning.
For each sample, we use different instruction prompt strategies, including general instruction (direct use), multi-modal instruction, in-context learning, and chain-of-thought prompting. Our trials show that LLaMA3 does not significantly benefit from prompt engineering. This is also observed during the fine-tuning stages, where performance worsens with in-context or chain-of-thought prompt strategies. Consequently, we opt to use general instruction for both fine-tuning and inference.

\textbf{Summary} Generally, GPT-4 performs better than LLaMA3 in most cases, although LLaMA3 sometimes achieves better performance after fine-tuning. Based on our manual review of the explanations provided by both models, GPT-4 offers more specific and accurate, or nearly accurate, descriptions compared to LLaMA3. We attribute this superiority to GPT-4's larger parameter scale, more diverse training datasets, and longer context window. Nevertheless, both models demonstrate great potential as effective time series anomaly detectors with explanations.



\section{Related Works for Time Series Anomaly Detection}
\label{app:more_related_work}
Traditional methods for detecting anomalies in time series data can be broadly categorized into several approaches. Prediction-based methods are the most prevalent, involving the training of a robust time series forecasting model, such as Prophet \citep{taylor2018forecasting} or the more recent transformer-based models like Informer \citep{zhou2021informer}. Anomalies are identified as points exhibiting significant deviations from forecasted values. Clustering-based approaches, exemplified by Isolation Forest (IForest) \citep{liu2008isolation}, utilize binary tree structures based on space partitioning, where nodes closer to the root are more likely to be anomalies. Pattern-matching approaches, such as Matrix Profile (MP) \citep{yeh2016matrix}, detect anomalies as subsequences with notably large nearest-neighbor distances. Reconstruction-based approaches, represented by Autoencoders \citep{sakurada2014anomaly}, learn to reconstruct data, flagging as outliers those points that significantly diverge from the reconstructed values.
The primary issue is that while most of them excel at capturing specific types of anomalies, they are also difficult to explain in terms of their detection results.

\section{Limitation and Future Work}
\label{app:limitations}
Although we have observed that GPT-4 can deliver good performance with minimal instructions, its current lack of public fine-tuning capabilities on GPT-4 prevents us from exploring whether fine-tuning on GPT-4 could achieve state-of-the-art (SOTA) performance. We are in the process of applying for access to the fine-tuning API for GPT-4 at the time of this submission. Therefore, in future work, we will explore the fine-tuning performance of GPT-4 if we gain access to this functionality.
Another limitation is that, while different approaches for representing time series in large language models (LLMs), such as images and embeddings, have been observed, we consider representing time series as pure text tokens in this research. One reason is based on our preliminary studies on GPT-4 with representing the time series as images only, we found that, although GPT-4 can grasp the overall shape of the time series, it experiences significantly more hallucinations in the indexes compared to representing the time series as pure tokens. 
Beyond the issue of hallucination in the indexes, another reason we avoided using embeddings and images to represent time series is the necessity for accurate identification and explanation of all anomaly points in time series anomaly detection tasks, where the accurate knowledge of the position and value of each time series point is essential. 
Another advantage of representing the time series as pure tokens is that it eliminates the need for data preprocessing, making it easier to integrate time series with textual prompts, thereby fully leveraging the LLMs' capabilities in handling tokens.
While we have not explored representing time series as embeddings or images in this research, we plan to conduct further analysis in this area to gain broader insights into using LLMs for time series anomaly detection tasks.

\end{document}